\documentclass[times,twocolumn,final]{elsarticle}
\usepackage{ipcai_arxiv}
\usepackage{framed,multirow}

\usepackage{amssymb}
\usepackage{latexsym}

\usepackage{url}
\usepackage{cite}
\usepackage{amsmath,amssymb,amsfonts,bm}
\usepackage{booktabs}
\usepackage{arydshln}
\usepackage{tabulary}
\usepackage{graphicx,epstopdf}
\usepackage{textcomp}
\usepackage{subcaption}
\usepackage{booktabs}
\usepackage[svgnames,table]{xcolor}
\usepackage{pifont}
\usepackage{arydshln}
\usepackage[colorlinks=true,citecolor=cyan,urlcolor=cyan]{hyperref}
\usepackage{graphicx}
\usepackage{placeins}
\usepackage{fancyhdr}
\usepackage{float}

\usepackage{listings}%
\usepackage{lipsum}
\usepackage{makecell}

\newcommand{\xmark}{\ding{55}}%

\fancypagestyle{firstpagestyle}{
    \fancyhf{} % Clear header and footer
    \fancyhead{} % Clear header
    \fancyfoot{} % Clear footer
    \fancyhead[CO]{\em \fontsize{9pt}{8pt}\selectfont This article has been accepted for publication in IEEE Transactions on Medical Imaging (TMI). DOI: 10.1109/TMI.2026.3756547}
}

\begin{document}

% \begin{frontmatter}

\title{Self-Supervised Uncalibrated Multi-View Video Anonymization in the Operating Room}

\author[1,2]{Keqi \snm{Chen}\corref{cor}}

\author[1,2,3]{Vinkle \snm{Srivastav}}

\author[2]{Armine \snm{Vardazaryan}}

\author[2]{Cindy \snm{Rolland}}

\author[2,4]{Didier \snm{Mutter}}

\author[1,2]{Nicolas \snm{Padoy}}

\cortext[cor]{Corresponding author: keqi.chen@unistra.fr}

\address[1]{University of Strasbourg, CNRS, INSERM, ICube, UMR7357, France}

\address[2]{IHU Strasbourg, 67000 Strasbourg, France}

\address[3]{Indian Institute of Technology (IIT) Madras, 600036 Chennai, India}

\address[4]{University Hospital of Strasbourg, 67000 Strasbourg, France}

\received{XXX}
\finalform{XXX}
\accepted{XXX}
\availableonline{XXX}
\communicated{XXX}

%%==================================%%
%% Sample for unstructured abstract %%
%%==================================%%
\begin{abstract}

Privacy preservation is a prerequisite for using video data in Operating Room (OR) research. Effective anonymization relies on the exhaustive localization of every individual; even a single missed detection necessitates extensive manual correction. However, existing approaches face two critical scalability bottlenecks: (1) they usually require manual annotations at each new clinical site for high accuracy; (2) multi-view approaches typically require camera calibration whenever cameras are repositioned. To address these problems, we propose a self-supervised multi-view video anonymization framework based on whole-body person detection and pose estimation, without manual annotations or camera calibration.
Our core strategy is to conduct self-supervised domain adaptation of a single-view detector by ``retrieving'' false negatives using temporal and multi-view context. We first apply an off-the-shelf whole-body detector with a low-score threshold to gather candidate detections. Then, we retrieve low-score false negatives that are consistent with high-score detections via tracking and self-supervised, uncalibrated multi-view association. The recovered detections serve as pseudo-labels to iteratively fine-tune the detector. Finally, we apply whole-body pose estimation on each detected person, and fine-tune the pose model using its own high-score predictions. Experiments on the simulated 4D-OR dataset and our real-surgery dataset achieve 99\% and 97\% recall, respectively. We further train a real-time whole-body detector using the generated pseudo-labels, achieving comparable performance and highlighting our method’s practical applicability. The code is available at \url{https://github.com/CAMMA-public/OR_anonymization}. 
\end{abstract}
%\end{frontmatter}

\maketitle
\thispagestyle{firstpagestyle}

\begin{figure}[t]
\centerline{\includegraphics[width=\columnwidth]{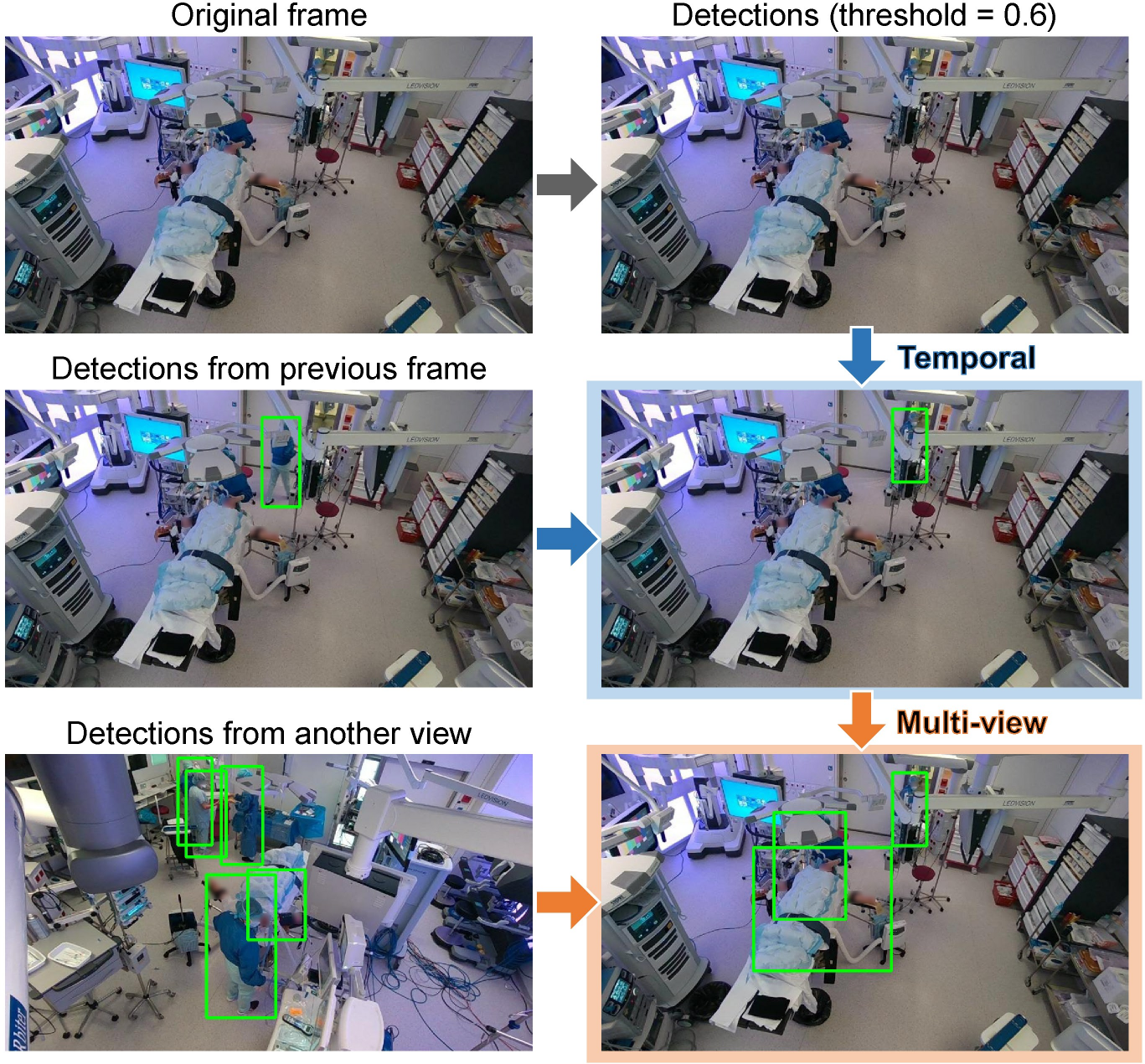}}
\caption{Examples of robust whole-body detection in the operating room using temporal and multi-view context. In the current frame, the detector fails to detect any instance. Using a tracker and detections from previous frame, we find one clinician; using multi-view person association and detections from another view, we find another clinician and the patient. }
\label{fig:intro}
\end{figure}

\section{Introduction}
\label{sec:introduction}

The operating room (OR) is a visually complex scene with various medical instruments and multiple clinicians. In recent years, hospitals across various sites have increasingly implemented video recording systems in the ORs with the overarching aim to develop novel context-aware systems~\citep{bastian2023disguisor,belagiannis2016parsing,czempiel2022surgical,hu2022multi,li2020robotic,ozsoy20224d,srivastav2018mvor}. These systems hold significant potential to detect adverse events in real time, optimize clinical workflow processes, facilitate safe human-robot collaboration, and assist in decision-making by automatically analyzing various clinical activities~\citep{maier2022surgical,mascagni2021or,vercauteren2019cai4cai,padoy2019machine}. However, due to the strict ethical regulations such as the Health Insurance Portability and Accountability Act (HIPAA)~\citep{act1996health} in the United States and the General Data Protection Regulation (GDPR)~\citep{regulation2018general} in the European Union, video anonymization is required to be applied on the recorded videos before external study in order to protect the privacy of both clinicians and patients. Specifically, it is usually required to blur the eyes, faces or even half bodies of all the persons in the OR. Since each surgery usually takes hours, it is unrealistic to manually anonymize these videos. Although generating low-resolution videos is an easy solution~\citep{srivastav2022unsupervised}, it leads to severe information loss of the holistic scene. Therefore, it is necessary to develop effective methods towards automatically localizing the persons in the OR for anonymization, in order to advance further OR-related study. 

A critical challenge in automatic OR video anonymization is to minimize missed detections, which are also known as false negatives. In practice, even a single missed detection may violate the ethical regulations, and thus a manual review process is strictly required to ensure that every person is properly anonymized through a frame-by-frame inspection. As the false negative rate increases, the time and effort of the review process will be significantly inflated since the reviewer will need to manually blur every neglected person. According to our experiments, although the state-of-the-art approach~\citep{carion2026sam} can anonymize over 92\% persons, there would still be more than 20,000 missed detections in a single one-hour video, which would require over 20 hours of manual correction assuming that it takes 4 seconds to detect and blur one missed person. Therefore, developing a robust detection framework that minimizes missed detections is essential to ensure the practical scalability of an anonymization pipeline. 

With the rapid development of object detection and human pose estimation, most existing works adopt pretrained detectors for OR video anonymization~\citep{flouty2018faceoff,issenhuth2019face}. However, these off-the-shelf detectors suffer from high false negative rates due to the substantial domain gap between the OR and natural images: clinicians typically wear masks and caps, and medical equipment causes significant occlusion. 
Although fine-tuning the models on domain-specific data can mitigate the domain gap, scaling such an approach is hindered by two critical bottlenecks. First, obtaining site-specific manual annotations is labor-intensive, and existing public datasets~\citep{srivastav2018mvor,ozsoy20224d} lack the diversity required for generalization. Second, while multi-view analysis~\citep{belagiannis2016parsing,srivastav2018mvor,ozsoy20224d,bastian2023disguisor,ozsoy2025mm} resolves single-view ambiguity, it traditionally relies on rigid camera calibration. 
Existing approaches fail to address these problems simultaneously: Issenhuth \emph{et al.}~\citep{issenhuth2019face} pioneer self-supervised domain adaptation using pseudo-labels but neglected temporal consistency and multi-view analysis; Bastian \emph{et al.}~\citep{bastian2023disguisor} achieve robustness via 3D point clouds but their work requires calibrated RGB-D setups that increase the deployment complexity. Therefore, developing a framework that is both self-supervised (label-free) and uncalibrated (setup-free) is essential for practical, scalable OR anonymization.

%Although fine-tuning the models with the OR data is effective, existing public OR datasets~\citep{srivastav2018mvor,ozsoy20224d} provide limited training samples in fixed clinical sites. It is time-consuming to manually collect and annotate enough data in every new clinical site. Moreover, the inherent single-view ambiguity makes accurate detection an ill-posed problem. Therefore, self-supervised domain adaptation~\citep{issenhuth2019face} and multi-view analysis~\citep{belagiannis2016parsing,srivastav2018mvor,ozsoy20224d,bastian2023disguisor,ozsoy2025mm} have become important research directions. Issenhuth \emph{et al.}~\citep{issenhuth2019face} demonstrate the potential of using high-score detections as pseudo-labels for domain adaptation, yet they fail to exploit temporal consistency in the videos. Bastian \emph{et al.}~\citep{bastian2023disguisor} propose to fuse the multi-view RGB and depth images into a 3d point cloud representation, and then develop an unsupervised approach to detect the keypoints and fit a 3d mesh for each person. However, their work requires setting up calibrated multi-view RGB-D cameras in the OR, which are challenging to acquire in practical OR settings. Therefore, developing a framework that is both self-supervised and capable of learning from uncalibrated RGB streams is essential for practical OR anonymization.

In this work, we propose a \emph{self-supervised uncalibrated multi-view video anonymization} approach, without using any annotations. We hypothesize that combining tracking and multi-view association will effectively reduce false negatives of an off-the-shelf detector through temporal and multi-view consistency. These augmented detections will serve as pseudo-labels to allow self-supervised domain adaptation.

Inspired by the observation in~\citep{issenhuth2019face} that the pose estimation-based approach performs better than face detection for coarse localization, we design a two-stage pipeline, where we first conduct whole-body bounding box detection in the OR to localize all the persons, and then apply whole-body pose estimation for each one of them. 
This pipeline offers two distinct advantages: (1) it ensures more robust localization in complex OR scenes through whole-body detection, which relies on more abundant features compared to face or visible-body detection; (2) it enables precise, granular anonymization of any desired body region (e.g., eyes, faces, or half bodies) according to different anonymization protocols. 

In our pipeline, the anonymization performance is heavily dependent on the recall of the first-stage whole-body detector, and thus we propose to reduce missed detections through tracking and multi-view association. Fig.~\ref{fig:intro} shows an example of our strategy: in the current frame, the detector fails to detect any person with a high score threshold; based on the detections from previous frame and a motion-based tracker, we find the clinician close to the door; based on the detections from a different view and multi-view person association, we find another clinician and the patient. 

%our core strategy focuses on minimizing missed detections by augmenting the whole-body detection with temporal and multi-view context. Specifically, we propose to utilize tracking and multi-view person association to retrieve false negative low-score detections, thereby enabling the generation of high-quality pseudo-labels for domain adaptation. Fig.~\ref{fig:intro} shows an example of our strategy: in the current frame, the detector fails to detect any person with high score threshold; based on the detections from previous frame and a motion-based tracker, we find the clinician close to the door; based on the detections from a different view and multi-view person association, we find another clinician and the patient. 

Our association strategy is inspired by a tracker named ByteTrack~\citep{zhang2022bytetrack}, which treats every detection box as a potential association match rather than considering only the high-score ones. Similarly, we extend it to the multi-view scenario. We first use an off-the-shelf whole-body detector to collect a large set of detections using a low-score threshold. Then, we apply ByteTrack~\citep{zhang2022bytetrack} in each view to retrieve the low-score boxes by tracking. Subsequently, using these tracked boxes in each view as queries,  we train a multi-view  association model in a self-supervised way~\citep{chen2025learning}, and conduct the association to retrieve the low-score detections in the other views that may have been neglected by tracking due to occlusion. In the end, we have all the retrieved detections as the augmented detections. 

To further refine the whole-body detection through domain adaptation, we use these augmented detections as pseudo-labels to fine-tune the whole-body detector, and repeat the detection, tracking, and multi-view association steps to generate better pseudo-labels iteratively. Our approach also enables real-time applicability as we can train a real-time detector using the generated high-quality pseudo-labels. After obtaining the final whole-body detections, we fine-tune the whole-body pose detector using its own high-score joint predictions, which can locate the desired keypoints more accurately. 

We evaluate our approach on both the 4D-OR dataset~\citep{ozsoy20224d} of simulated surgeries and our collected dataset of real surgeries in a more complex scene. For each person, we annotate (1) the whole-body bounding box, (2) the ``hard case'' flag (over 67\% occlusion), and (3) three keypoints (eyes and chin) if they are visible. 
We also apply two strict metrics to quantify robustness against challenging scenes: (1) hard-case recall, which calculates the recall rate of the hard cases; (2) holistic recall, which measures the percentage of subjects anonymized across all camera views in which they appear~\citep{bastian2023disguisor}. 
Experimental results show that our approach outperforms existing methods, specifically by achieving 99\% recall on 4D-OR and 97\% recall on our dataset. 
Compared to state-of-the-art approaches, ours reduces 10,000 missed detections in a one-hour real surgical video, which saves significant manual correction time.

We summarize our contributions as follows:
\begin{itemize}
    \item We address video anonymization in the OR without using any annotations.
    \item We propose a two-stage anonymization pipeline that detects whole bodies and estimates whole-body keypoints, which achieves state-of-the-art performance and supports anonymization of any desired body region.
    \item We propose to combine tracking and multi-view person association for more robust whole-body detection using temporal and multi-view context.
    \item We solve domain adaptation and real-time detection by using spatial-temporally augmented detections as pseudo-labels for model fine-tuning.
    \item We use strict metrics including hard-case recall and holistic recall, and evaluate different approaches on both actor-simulated and real surgical videos at three anonymization levels: whole-body, face, and eye.
\end{itemize}

\section{Related work}
\label{sec:related}
\subsection{Operating room datasets}
Despite the growing number of intra-corporeal OR datasets in recent years~\citep{nwoye2023cholectriplet2022,murali2023endoscapes,lavanchy2024challenges,che2026lemon}, room-level OR datasets captured by ceiling-mounted cameras remain scarce. Belagiannis \emph{et al.}~\citep{belagiannis2016parsing} propose the first multi-view OR dataset named MultiHumanOR, which consists of simulated surgeries with human pose annotations. To introduce data captured during real interventions, Srivastav \emph{et al.}~\citep{srivastav2018mvor} propose the MVOR image dataset, the first multi-view RGB-D dataset with 3d human poses. In order to untangle the interactions between clinicians and objects, Özsoy \emph{et al.}~\citep{ozsoy20224d} propose the 4D-OR dataset of simulated knee surgeries with semantic scene graph annotations. Then, to enable multi-modal analysis in the OR, they further propose the MM-OR dataset~\citep{ozsoy2025mm}, which consists of robotic knee replacement surgeries with a wide range of data sources. With these existing datasets, several computer vision tasks in the OR are supported, such as human pose estimation~\citep{hansen2019fusing,srivastav2022unsupervised}, semantic scene graph generation~\citep{ozsoy2023labrad,pei2024s2former}, surgical phase recognition~\citep{ozsoy2024holistic}, and panoptic segmentation~\citep{ozsoy2025mm}. However, neither 4D-OR nor MM-OR captures real surgical procedures that are much more visually complex. Consequently, they fail to provide the necessary context for addressing missed detections in real-world video anonymization.

\subsection{Video anonymization in the operating room}

Due to the explosive growth of video data and rising concerns over personal privacy, video anonymization has received growing attention in recent years, where face anonymization plays a significant role~\citep{gafni2019live,maximov2020ciagan,rosberg2023fiva}. In the OR, anonymizing recorded videos is a critical prerequisite for external studies to protect the identities of both clinicians and patients. Although generating low-resolution videos~\citep{srivastav2022unsupervised} or using depth videos~\citep{jamal2022multi} are feasible solutions in certain scenarios, the inevitable information loss brings severe challenges to the subsequent video analysis. Therefore, face anonymization in the OR becomes an important research area, which relies on accurate face detection. However, the advanced face detection models~\citep{najibi2017ssh,hu2017finding} trained on the WIDER Faces dataset~\citep{yang2016wider} perform poorly in the complex OR scene due to the large domain gap~\citep{issenhuth2019face}. To address the problem, Flouty \emph{et al.}~\citep{flouty2018faceoff} propose an OR dataset named FaceOff for model fine-tuning, which contains 6,371 images with face annotations. To implement domain adaptation without annotations, Issenhuth \emph{et al.}~\citep{issenhuth2019face} propose an iterative self-supervised learning strategy, where they select high-score detections as pseudo-labels in each round to fine-tune the face detector iteratively. However, this approach still suffers from the severe occlusion and does not take advantage of temporal context. To address the single-view ambiguity, Bastian \emph{et al.}~\citep{bastian2023disguisor} utilize multi-view RGB-D cameras to do 3d human pose estimation, but it requires accurate camera poses that are hard to obtain. In contrast, our approach utilizes both temporal and multi-view complementary information for better person localization in the OR, without using any annotations or camera calibration.

\subsection{Tracking and multi-view person association}

Despite the rapid development of image-based object detection, the complexity and occlusions in the OR scene remain significant challenges, leading to high false negative rates. To address this deficiency, it is necessary to integrate additional temporal and multi-view information. Specifically, multi-object tracking can address transient occlusions, while multi-view association can recover objects obscured in one view but visible in the others. 

The advanced multi-object trackers are mostly tracking-by-detection, where they use appearance~\citep{wojke2017simple} and motion features~\citep{kalman1960new} to associate the detected boxes with the activated tracklets. To enhance robustness against occlusion, ByteTrack~\citep{zhang2022bytetrack} proposes a strategy that includes all the detections regardless of scores in the candidate pool, in order to prevent valid targets from being discarded. This approach has since become a widely-adopted paradigm~\citep{aharon2022bot,maggiolino2023deep,stanojevic2024boosttrack}. In our work, we also use a motion-based tracker to retrieve low-score boxes that exhibit temporal consistency with activated tracklets.

Similarly with tracking, multi-view person association aims to associate the detection boxes in synchronized camera views using appearance~\citep{vo2020self,gan2021self} or geometric features~\citep{luna2022graph,seo2023vit}. In order to encode both appearance and geometric features for robust multi-view association without using any annotations or camera calibration, Self-MVA~\citep{chen2025learning} trains a multi-view encoder by distinguishing whether two images from different camera views are captured at the same time. In this work, we optimize Self-MVA to specifically address false negatives: instead of finding associations between high-score boxes, we expand the search space to include all the detected boxes in the other views regardless of scores. This allows us to retrieve the heavily occluded persons if they are visible in the other views. 

\begin{figure*}[!t]
\centerline{\includegraphics[width=0.85\textwidth]{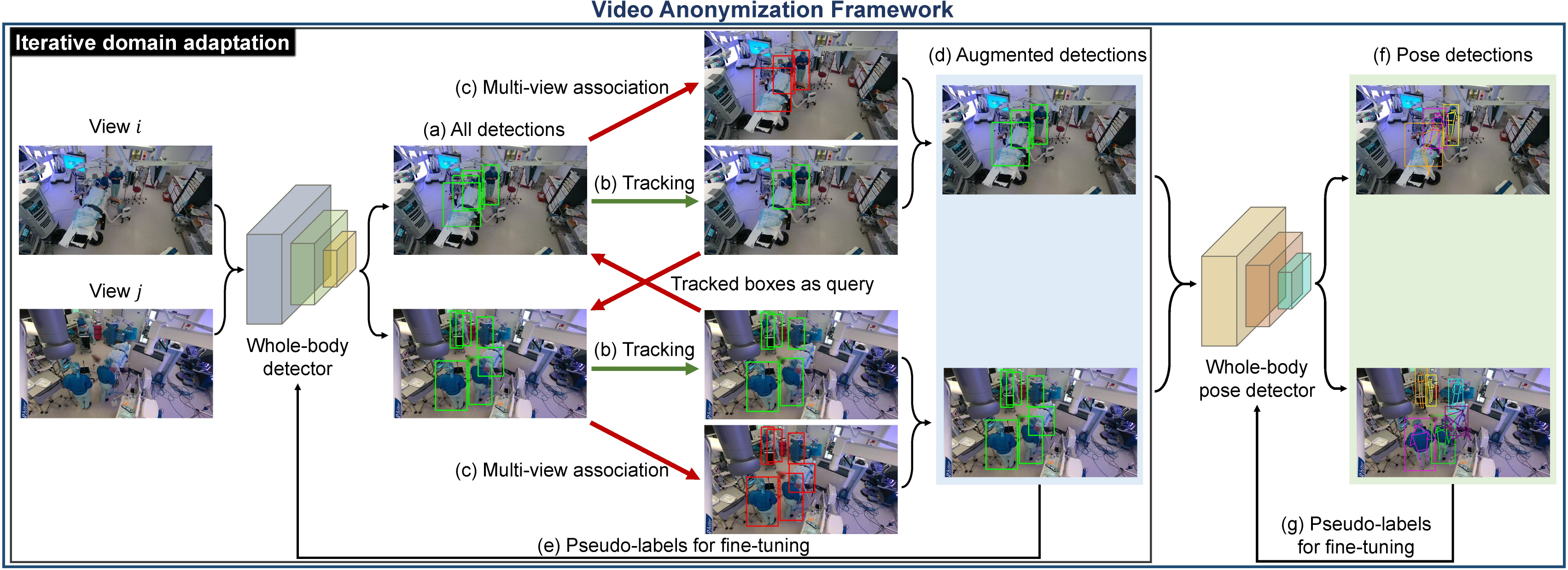}}
\caption{Framework of our two-stage approach, including whole-body detection and whole-body pose estimation. During training, we apply iterative domain adaptation for whole-body detection: in each round, (a) we use a whole-body detector to generate detections by applying a low-score threshold; (b) we use tracking to obtain tracklets; (c) we use tracklets and multi-view association to find missing detections; (d) we merge tracklets and association results as augmented detections; (e) we fine-tune the detector using augmented detections as pseudo-labels. For the whole-body pose detector, we (f) predict poses and (g) fine-tune the model using the high-score predictions on the augmented whole-body detections.}
\label{fig:method}
\end{figure*}

\section{Methodology}
\label{sec:methodology}

\subsection{Problem overview}

With fixed camera positions in the OR, given multi-view videos of $n$ consecutive RGB frames $\mathcal{V} = \left\{ x_1, x_2, ..., x_n \right\}$ where $ x_i \in \mathbb{R}^{C \times 3 \times H \times W}$ represents a multi-view image set from $C$ cameras with height $H$ and width $W$, the goal is to conduct video anonymization by detecting sensitive anatomical regions, such as faces, eyes, and half bodies. To address the problem, we propose a two-stage pipeline, where we first detect the whole bodies using temporal and multi-view context, and then conduct whole-body pose estimation to localize the keypoints. 

In the following, we describe our framework as shown in Fig.~\ref{fig:method}. We first use an off-the-shelf whole-body detector~\citep{zheng2022progressive} to initially collect a large set of detections in each view with a low-score threshold. Then, to recover false negatives, we formulate both tracking and multi-view association as retrieval tasks: the high-score detections serve as queries, while the complete set of detections in the next frame or in the other views constitutes the candidate pool (referred to as the ``gallery''). Specifically, we apply a tracker~\citep{zhang2022bytetrack} in each view to obtain all the tracklets. Using these tracklets as queries and all the detected boxes in the other views as the gallery, we conduct multi-view person association by training a geometric encoder in a self-supervised way~\citep{chen2025learning}, which further retrieves the false negatives using multi-view geometry. In the end, we have the augmented whole-body detections. In order to implement domain adaptation, we use these augmented detections as pseudo-labels, and fine-tune the whole-body detector using OR data. By repeating the previous detection, tracking, and multi-view association steps, we iteratively generate new augmented detections with higher quality, which also supports the training of real-time detectors. Finally, we conduct whole-body pose estimation, and fine-tune the model using its own high-score joint predictions. During inference, we blur the sensitive regions for each person. We introduce each key step in detail as follows.

\begin{figure}[t]
\centerline{\includegraphics[width=\columnwidth]{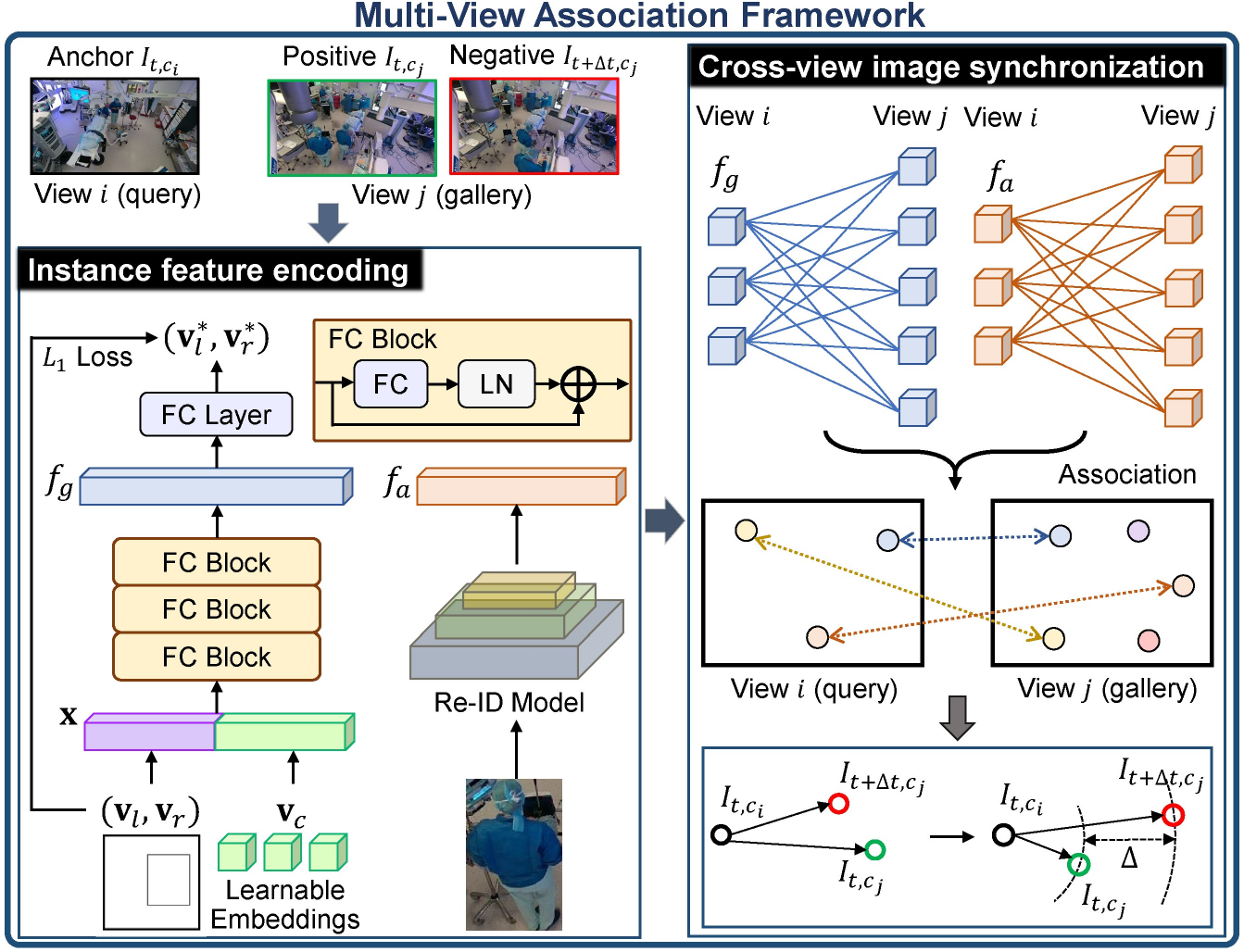}}
\caption{Framework of the self-supervised multi-view association approach~\citep{chen2025learning} (FC = Fully-Connected, LN = Layer Normalization, Re-ID = Re-IDentification). For each anchor image, we construct a triplet by selecting synchronized and non-synchronized images in a different view. Then, we encode each detected person's appearance features using a person Re-ID model, map their geometric information to a unified geometric feature space using positional encodings and learnable camera embeddings, and then decode the original 2d position. Finally, we use encoded features to conduct instance association along with triplet-based metric learning. Best viewed in color. }
\label{fig:self-mva}
\end{figure}

\subsection{Whole-body detection and tracking}

In order to circumvent the dependence on the ground-truth data, we first utilize an off-the-shelf detector~\citep{zheng2022progressive} trained on the CrowdHuman dataset~\citep{shao2018crowdhuman} to detect the whole bodies. Compared to face detection or visible-body detection, whole-body detection takes more advantage of the partially visible body features, and possesses the ability to deduce the positions of the whole bodies even when faced with severe occlusion. After applying the whole-body detector on the videos, we have all the detected boxes as $\mathcal{D}$. 

Due to the intrinsic domain gap between the OR and natural scenes, the high-score detections tend to ignore the persons that are partially occluded, while the low-score ones contain too many false positives. To retrieve the false negatives using temporal context, we apply ByteTrack~\citep{zhang2022bytetrack} for each video in both forward and backward temporal directions. Specifically, we classify all detections $\mathcal{D}$ into high-score set $\mathcal{D}^h$ and low-score set $\mathcal{D}^l$ using a manually-defined threshold. For each timestamp $t$, we first use the Kalman Filter to predict the locations of the activated tracklets (tracklets successfully associated with prior detections), and then associate them with the high-score boxes $\mathcal{D}_t^h$, using IoU distances and Hungarian matching~\citep{kuhn1955hungarian}. Then, for the unmatched tracklets, we re-associate them with the low-score boxes $\mathcal{D}_t^l$ to find potential false negatives. Lastly, for the remaining high-score boxes, we initialize them as new tracklets; for the tracklets that are not associated for some time, we de-activate them. After processing the videos, we have the tracked boxes as $\mathcal{T}$, which is a superset of $\mathcal{D}^h$, as shown in Fig.~\ref{fig:method}.

\subsection{Self-supervised multi-view person association}

To further retrieve the false negatives using multi-view geometry, 
we adapt the Self-MVA~\citep{chen2025learning} framework, a self-supervised approach for uncalibrated multi-view person association. Self-MVA learns a geometric encoder through the pretext task of cross-view image synchronization, which aims to distinguish whether two images of different views are captured at the same time. This task is solved by encoding instance features and conducting cross-view instance association to compute the image-level distance. In order to retrieve the false negatives, we extend the original association procedure by expanding its search space. While Self-MVA only associates high-score boxes, we expand the search space to include all the detected boxes in the other views regardless of scores. This modification enables the encoder to associate high-score detections with geometrically consistent low-score candidates, allowing detections that would otherwise be discarded to be recovered.
% we propose to train a geometric encoder that measures boxes' geometric distances in different views. Inspired by Self-MVA~\citep{chen2025learning}, a self-supervised uncalibrated multi-view association approach, we train the encoder via a pretext task, cross-view image synchronization, which aims to distinguish whether two images of different views are captured at the same time. Specifically, we solve this task by encoding instance features and conducting cross-view instance association to compute the image-level distance. Unlike the original Self-MVA that only associates high-score boxes, we expand the search space to include all the detected boxes in the other views regardless of scores. Therefore, the trained encoder is capable of finding geometric associations between high-score and low-score boxes, and thus enables retrieving missed detections. 
The learning framework is shown in Fig.~\ref{fig:self-mva}.

For each detected person with normalized top-left and bottom-right corner point positions $(\mathbf{v}_l, \mathbf{v}_r)$, we first map them to a higher dimensional hypersphere~\citep{tancik2020fourier} as follows, where $\mathbf{b}_j (1 \leq j \leq N)$ are the learnable Fourier basic frequencies:
\begin{equation}
\gamma(\mathbf{v}) = [\sin \mathbf{f}_1, \cos \mathbf{f}_1, ...,
\sin \mathbf{f}_N, \cos \mathbf{f}_N]^{\mathrm{T}},
 \label{eq:pos_encoding1}
\end{equation}
\begin{equation}
\mathbf{f}_j = 2\pi \mathbf{b}_j^{\mathrm{T}}\mathbf{v}.
 \label{eq:pos_encoding2}
\end{equation}

Then, to encode camera poses, we maintain learnable camera embeddings $\mathbf{v}_c \in \mathbb{R}^{C \times V}$ of size $V$ for $C$ views. For each person, we concatenate its 2d box representation and the camera embedding and pass it through fully-connected (FC) blocks $\mathrm\mathtt{F}(\cdot)$ to obtain the geometric features $f_{g}$:
\begin{equation}
\mathbf{x} = [\gamma(\mathbf{v}_l), \gamma(\mathbf{v}_r), \mathbf{v}_c],
 \label{eq:pos_in}
\end{equation}
\begin{equation}
f_{g} = \mathrm\mathtt{F}(\mathbf{x}).
 \label{eq:pos_out}
\end{equation}
Consequently, for all detections $\mathcal{D}$, we obtain their geometric features $\mathcal{F}_g=\left\{ f^1_{g}, ..., f^{|\mathcal{D}|}_{g} \right\}$. Additionally, using an off-the-shelf person re-identification model~\citep{zhou2021learning}, we obtain the appearance features $\mathcal{F}_a=\left\{ f^1_{a}, ..., f^{|\mathcal{D}|}_{a} \right\}$.

To compute the overall distance between two images, we apply Hungarian matching to bridge the gap between instance-wise and image-wise distances. Inspired by ByteTrack~\citep{zhang2022bytetrack} that associates tracklets with all the boxes regardless of scores, we extend it to multi-view domain, by using tracked boxes $\mathcal{T}_t$ at time $t$ as queries, and using all the detected boxes $\mathcal{D}_t$ as the gallery for association. Specifically, given an anchor image $I_{t,1}$ and another image $I_{t,2}$, we use $\mathcal{T}_{t,1}$ as queries and $\mathcal{D}_{t,2}$ as the gallery. Then, we compute the normalized appearance and geometric Euclidean distance matrix $(\mathcal{E}_a, \mathcal{E}_g)\in \mathbb{R}^{|\mathcal{T}_{t,1}| \times |\mathcal{D}_{t,2}|}$, and take their weighted sum as the overall distance matrix $\mathcal{D}$, representing instance-wise distances:
\begin{equation}
\mathcal{E} = \alpha \mathcal{E}_a + (1 - \alpha) \mathcal{E}_g.
 \label{eq:dis}
\end{equation}
After applying Hungarian matching, we obtain the matched row and column indices $(\mathcal{M}_r, \mathcal{M}_c)\in \mathbb{R}^{m}$ of matrix $\mathcal{E}$, where $m=\min(|\mathcal{T}_{t,1}|,|\mathcal{D}_{t,2}|)$. In the end, we average the pairwise instance-wise distances as the image-wise distance $\mathcal{H}$: 
\begin{equation}
\mathcal{H} = \frac{1}{m}\sum \mathcal{E}[\mathcal{M}_r; \mathcal{M}_c].
 \label{eq:img_dis}
\end{equation}

Finally, we conduct triplet-based metric learning. Given an anchor image $I_{t, c_i}$ from view $i$ at time $t$, we construct a triplet $(I_{t, c_i}, I_{t, c_j}, I_{t+\Delta t, c_j})$, where $(I_{t, c_i}, I_{t, c_j})$ is the positive pair, and $(I_{t, c_i}, I_{t+\Delta t, c_j})$ is the negative pair. $\Delta t$ is randomly chosen that satisfies $\Delta t \in [t_{\mathrm{min}},t_{\mathrm{max}}]$, where $[t_{\mathrm{min}},t_{\mathrm{max}}]$ is manually defined frame range. Then we compute the triplet loss $L_{\mathrm{syn}}$ for training:
\begin{equation}
L_{\mathrm{syn}}=\max (0, \mathcal{H}_{I_{t, c_i}, I_{t, c_j}} - \mathcal{H}_{I_{t, c_i}, I_{t+\Delta t, c_j}} + \triangle),
 \label{eq:triplet_syn}
\end{equation}
where $\triangle$ is the margin between positive and negative pairs. 

\begin{figure}[t]
\centerline{\includegraphics[width=\columnwidth]{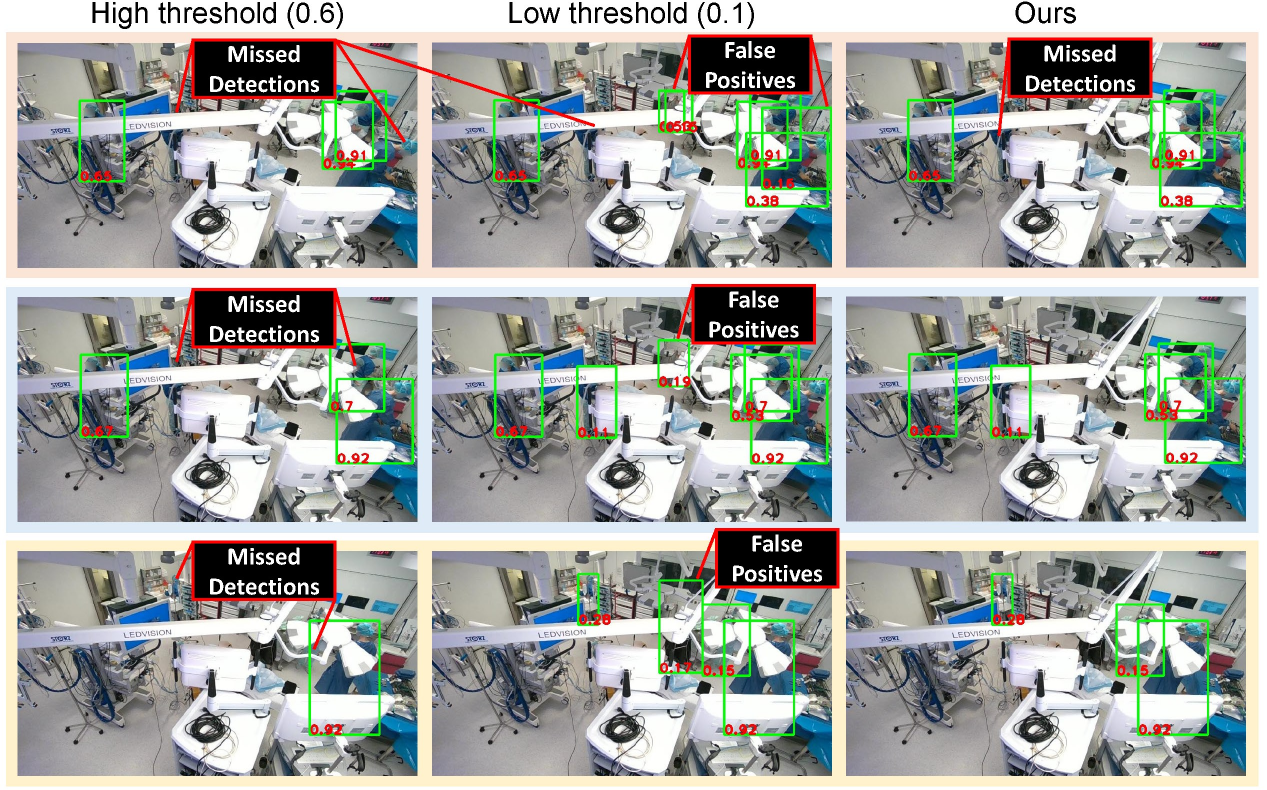}}
\caption{Comparison of different ways to generate pseudo-labels: (1) setting the score threshold to 0.6 will introduce false negatives (missed detections); (2) setting the score threshold to 0.1 will introduce false positives; (3) our strategy combining tracking and multi-view person association obtains pseudo-labels of the best quality. }
\label{fig:labels}
\end{figure}

To reduce the solution space and stabilize the training, we apply self-supervised multi-view re-projection loss as introduced in~\citep{chen2025learning}. In order to prevent potential information loss during geometric feature encoding, we project $f_{g}$ back to the original view using a single linear layer, obtaining the estimated bounding boxes $(\mathbf{v}_l^{\ast}, \mathbf{v}_r^{\ast})$, as shown in Fig.~\ref{fig:self-mva}. Then, we use $(\mathbf{v}_l, \mathbf{v}_r)$ as the pseudo-labels, and compute the $L_1$ loss as the re-projection loss $L_{\mathrm{pro}}$:
\begin{equation}
L_{\mathrm{pro}}=\mathcal{L}_1(\mathbf{v}_l^{\ast}, \mathbf{v}_l) + \mathcal{L}_1(\mathbf{v}_r^{\ast}, \mathbf{v}_r).
 \label{eq:loss_loc}
\end{equation}

In general, the final loss function is as follows: 
\begin{equation}
L=L_{\mathrm{syn}} + L_{\mathrm{pro}}.
 \label{eq:loss}
\end{equation}

During inference, for each multi-view image set $\left\{ I_{1}, I_{2}, ..., I_{C} \right\}$, we have tracked boxes $\left\{\mathcal{T}_{1}, \mathcal{T}_{2}, ..., \mathcal{T}_{C} \right\}$ and all the detected boxes $\left\{\mathcal{D}_{1}, \mathcal{D}_{2}, ..., \mathcal{D}_{C} \right\}$. We treat each tracked box in each image as the anchor, and conduct person association with the detected boxes in the remaining $C-1$ images. Consequently, we obtain the associated boxes $\left\{\mathcal{V}_{1}, \mathcal{V}_{2}, ..., \mathcal{V}_{C} \right\}$. After merging $\mathcal{T}$ and $\mathcal{V}$ using non-maximum suppression (NMS), we have the temporally and spatially associated boxes $\mathcal{A}$ as the augmented detections: 
\begin{equation}
\mathcal{A}=\mathrm{NMS}(\mathcal{T}, \mathcal{V}).
 \label{eq:final_boxes}
\end{equation}

\begin{table*}[h]
\scriptsize
\centering
\caption{Comparison with existing approaches on our dataset of real surgeries and 4D-OR dataset of simulated surgeries. If not specified, we use P-D-DETR as the base detector for self-supervised approaches. \textbf{Best scores}, \underline{second best scores}.}\label{tab:results}
\begin{tabular}{c|cccc|ccccc|ccccc}
\toprule%
\multirow{3}{*}{Method} & \multicolumn{14}{c}{Our dataset of real surgeries} \\ 
\cmidrule{2-15}
& \multicolumn{4}{c|}{Whole body} & \multicolumn{5}{c|}{Face}  & \multicolumn{5}{c}{Eye} \\
\cmidrule{2-15}
& F3 & P & R & $\text{R}_{\mathrm{hard}}$ & F3 & P & R & $\text{R}_{\mathrm{hard}}$ & HoR & F3 & P & R & $\text{R}_{\mathrm{hard}}$ & HoR \\
\midrule
RetinaFace~\citep{deng2020retinaface} & - & - & - & - & 41.13 & 72.13 & 39.25 & 8.91 & 22.87 & 40.74 & 71.76 & 38.87 & 8.91 & 22.62 \\
Head-YOLOv13~\citep{lei2025yolov13} & - & - & - & - & 42.05 & 36.76 & 42.73 & 12.63 & 27.26 & 32.37 & 28.30 & 32.89 & 12.10 & 20.41 \\
SAM3 (face)~\citep{carion2026sam} & - & - & - & - & 89.56 & 69.12 & 92.61 & 76.73 & 81.32 & 88.16 & 63.87 & 92.04 & 78.46 & 79.65 \\
SAM3 (head)~\citep{carion2026sam} & - & - & - & - & 84.83 & 52.71 & 90.99 & \textbf{86.04} & 77.67 & 86.07 & 60.51 & 90.30 & \textbf{83.64} & 73.31 \\
SAM3 (person)~\citep{carion2026sam} & 59.40 & 38.25 & 63.29 & 44.90 & 93.74 & 74.94 & 96.43 & 78.72 & 93.47 & 93.61 & 76.51 & 95.99 & 79.12 & 93.19 \\
P-D-DETR~\citep{zheng2022progressive} & 84.42 & 65.25 & 87.27 & 66.55 & 92.35 & 74.80 & 94.82 & 73.54 & 91.14 & 92.29 & 76.07 & 94.53 & 72.87 & 90.98 \\
Mean Teacher~\citep{hao2024simplifying} & 86.04 & 73.80 & 87.66 & 66.75 & 92.85 & \underline{78.12} & 94.84 & 73.94 & 91.32 & 92.74 & 78.67 & 94.62 & 74.60 & 91.04 \\
Iter-Score~\citep{issenhuth2019face} & 83.86 & \textbf{95.76} & 82.72 & 51.19 & 92.50 & \textbf{80.15} & 94.11 & 67.02 & 90.28 & 92.34 & \textbf{80.77} & 93.84 & 66.09 & 89.97 \\
Iter-Score (from SAM3) & 52.61 & 50.18 & 52.89 & 24.03 & 93.02 & 78.03 & 95.05 & 71.28 & 91.61 & 92.49 & \underline{79.98} & 94.13 & 68.48 & 90.25 \\
\midrule
Ours (P-D-DETR) & \textbf{94.84} & \underline{85.86} & \textbf{95.95} & \textbf{88.98} & \textbf{94.30} & 75.55 & \underline{96.97} & 82.85 & \textbf{94.92} & \underline{94.22} & 77.35 & \underline{96.56} & 81.25 & \textbf{94.20} \\
Ours (DEIM~\citep{huang2025deim}) & \underline{90.11} & 83.44 & \underline{90.91} & \underline{77.63} & \underline{94.19} & 74.10 & \textbf{97.12} & \underline{83.91} & \textbf{94.92} & \textbf{94.24} & 76.88 & \textbf{96.67} & \underline{81.78} & \underline{94.16} \\
Ours (DEIM, from SAM3) & 57.12 & 40.29 & 59.90 & 42.35 & 94.07 & 74.69 & 96.86 & 81.65 & \underline{94.42} & 93.81 & 76.81 & 96.18 & 80.32 & 93.50 \\
\midrule
\multirow{3}{*}{Method} & \multicolumn{14}{c}{4D-OR dataset of simulated surgeries} \\ 
\cmidrule{2-15}
& \multicolumn{4}{c|}{Whole body} & \multicolumn{5}{c|}{Face}  & \multicolumn{5}{c}{Eye} \\
\cmidrule{2-15}
& F3 & P & R & $\text{R}_{\mathrm{hard}}$ & F3 & P & R & $\text{R}_{\mathrm{hard}}$ & HoR & F3 & P & R & $\text{R}_{\mathrm{hard}}$ & HoR \\
\midrule
RetinaFace~\citep{deng2020retinaface} & - & - & - & - & 72.69 & 79.35 & 72.01 & 49.10 & 41.86 & 72.66 & 79.30 & 71.99 & 49.10 & 42.82 \\
Head-YOLOv13~\citep{lei2025yolov13} & - & - & - & - & 53.14 & 57.14 & 52.73 & 25.63 & 20.48 & 42.46 & 46.24 & 42.08 & 43.32 & 16.08 \\
SAM3 (face)~\citep{carion2026sam} & - & - & - & - & 97.65 & 86.23 & 99.11 & 94.58 & 92.37 & 97.53 & 86.13 & 98.99 & 94.58 & 95.95 \\
SAM3 (head)~\citep{carion2026sam} & - & - & - & - & 90.12 & 55.21 & 96.93 & \underline{96.39} & 83.02 & 91.35 & 62.41 & 96.32 & \underline{96.03} & 65.70 \\
SAM3 (person)~\citep{carion2026sam} & 79.26 & 55.32 & 83.27 & 51.25 & \underline{98.26} & 90.44 & \underline{99.21} & 94.58 & \underline{98.01} & \underline{98.29} & 91.49 & \underline{99.11} & 95.31 & \underline{97.94} \\
P-D-DETR~\citep{zheng2022progressive} & 82.62 & 71.37 & 84.09 & 47.99 & 96.45 & 89.35 & 97.31 & 73.65 & 95.81 & 96.63 & 90.14 & 97.40 & 75.81 & 96.98 \\
Mean Teacher~\citep{hao2024simplifying} & 83.93 & 76.63 & 84.83 & 48.40 & 96.69 & 89.60 & 97.55 & 76.90 & 96.77 & 96.90 & 90.25 & 97.70 & 81.23 & 97.53 \\
Iter-Score~\citep{issenhuth2019face} & 81.94 & \textbf{91.79} & 80.98 & 30.21 & 93.35 & 91.01 & 93.62 & 55.23 & 86.46 & 93.40 & \underline{91.63} & 93.60 & 56.32 & 86.94 \\
Iter-Score (from SAM3) & 75.90 & 73.98 & 76.11 & 28.26 & 97.28 & \textbf{92.07} & 97.90 & 75.45 & 96.63 & 97.31 & \textbf{92.69} & 97.85 & 76.90 & 97.53 \\
\midrule
% Ours (P-D-DETR, 8.63 FPS) & 79.90 & \textbf{90.55} & \textbf{88.41} & \textbf{90.55} & 82.84 & \textbf{98.59} & \textbf{90.57} & \underline{96.75} & \underline{97.18} & 83.17 & 97.35 & \textbf{90.69} & \underline{94.58} & \underline{96.49} \\
% Ours (DEIM~\citep{huang2025deim}, 52.09 FPS) & \underline{82.68} & \underline{90.08} & \textbf{88.41} & \underline{90.08} & \underline{86.37} & 98.42 & \underline{90.60} & 96.75 & 97.32 & \underline{86.94} & \textbf{97.90} & \underline{90.70} & \textbf{94.95} & \textbf{96.77} \\
Ours (P-D-DETR) & \underline{88.04} & 77.28 & \underline{89.43} & \underline{60.21} & 97.21 & \underline{91.62} & 97.87 & 77.26 & 96.77 & 97.73 & 90.43 & 98.62 & 89.89 & 97.66 \\
Ours (DEIM~\citep{huang2025deim}) & \textbf{89.25} & \underline{81.02} & \textbf{90.26} & \textbf{66.18} & 97.04 & 90.10 & 97.87 & 78.70 & 97.39 & 97.66 & 90.61 & 98.52 & 89.17 & 97.46 \\
Ours (DEIM, from SAM3) & 78.67 & 55.61 & 82.48 & 54.24 & \textbf{98.53} & 89.64 & \textbf{99.63} & \textbf{97.83} & \textbf{99.31} & \textbf{98.52} & 90.49 & \textbf{99.51} & \textbf{98.19} & \textbf{99.04} \\ 
\bottomrule
\end{tabular}
\end{table*}

\subsection{Iterative fine-tuning and real-time testing}

By integrating tracking and multi-view person association into the framework, our approach can recover a significant number of false negative detections. These augmented detections not only establish a more robust baseline for anonymization, but can also serve as high-quality pseudo-labels for fine-tuning the detector, thereby enabling effective domain adaptation to the challenging OR environment. As shown in Fig.~\ref{fig:labels}, compared to setting different score thresholds, our strategy obtains much better pseudo-labels. 

During training, we use $\mathcal{A}$ as pseudo-labels, along with the CrowdHuman dataset~\citep{shao2018crowdhuman}, to fine-tune the original detector with data in both OR and natural scenes. After model fine-tuning, we re-conduct the detection, tracking and multi-view association, and generate new augmented detections with higher quality. By repeating the above steps iteratively, we continuously improve the detector and retrieve more persons in the OR. Furthermore, we also use pseudo-labels to train a real-time detector~\citep{huang2025deim}, which enables real-time testing in real-world applications, with negligible impact on performance.

\subsection{Whole-body pose estimation}

After obtaining whole-body detections, we apply whole-body pose estimation for each person that detects 133 keypoints, which enables anonymization. 
To achieve self-supervised domain adaptation, we adopt the strategy from~\citep{issenhuth2019face} to fine-tune a state-of-the-art model using its own joint predictions. 
% To achieve self-supervised domain adaptation, we propose to fine-tune a state-of-the-art model using its own joint predictions, similar to~\citep{issenhuth2019face}. 
Specifically, we use an RTMPose model~\citep{jiang2023rtmpose} trained on 14 public datasets
%~\citep{wu2017ai,li2018crowdpose,andriluka14cvpr,cai2020learning,li2020pastanet,andriluka2018posetrack,jin2020whole,lin2023one,ju2023humanart,wayne2018lab,sagonas2013300,burgos2013robust,liu2020new,Moon_2020_ECCV_InterHand2.6M} 
to estimate initial keypoints of the augmented whole-body detections $\mathcal{A}$. Then, we only keep the high-score keypoints as the pseudo-labels. Finally, we fine-tune the model using the OR data and the COCO-Wholebody dataset~\citep{jin2020whole} simultaneously, which allows better performance.

\begin{figure*}[!t]
\centerline{\includegraphics[width=0.9\textwidth]{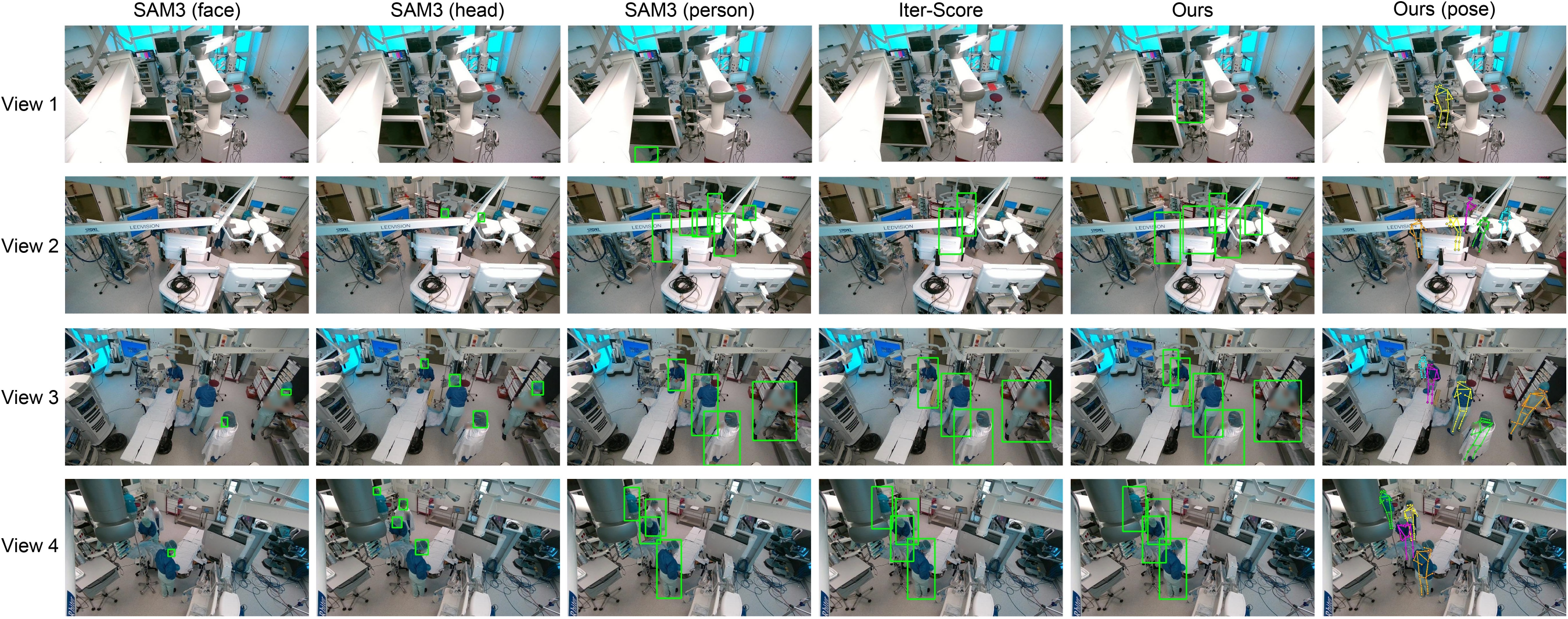}}
\caption{Examples of different video anonymization approaches on our dataset of real surgeries. Compared to SAM3 (face)~\citep{carion2026sam} or SAM3 (head), SAM3 (person) sets up a stronger baseline, but still fails in challenging cases. Iter-Score~\citep{issenhuth2019face} regresses the bounding boxes more accurately, but fails to retrieve the false negatives. Our approach detects all the persons using temporal and multi-view context, and supports flexible anonymization with human pose estimation.}
\label{fig:results}
\end{figure*}

\begin{table*}[h]
\scriptsize
\centering
\caption{Video-level test results on our dataset of real surgeries.}\label{tab:single_video_test}
\begin{tabular}{l|c|cccc|ccccc|ccccc}
\toprule%
& \multirow{2}{*}{Method} & \multicolumn{4}{c|}{Whole body} & \multicolumn{5}{c|}{Face} & \multicolumn{5}{c}{Eye} \\
\cmidrule{3-16}
& & F3 & P & R & $\text{R}_{\mathrm{hard}}$ & F3 & P & R & $\text{R}_{\mathrm{hard}}$ & HoR & F3 & P & R & $\text{R}_{\mathrm{hard}}$ & HoR \\
\midrule
{\multirow{2}{*}{{\makecell{Video 1}}}} & SAM3 (person) & 66.27 & 45.15 & 69.90 & 50.35 & 94.88 & 75.19 & \textbf{97.72} & 80.36 & \textbf{95.95} & 94.70 & 76.67 & 97.24 & 80.06 & 95.14 \\
& Ours (P-D-DETR) & \textbf{94.47} & \textbf{87.79} & \textbf{95.28} & \textbf{85.32} & \textbf{95.22} & \textbf{77.37} & \textbf{97.72} & \textbf{83.38} & \textbf{95.95} & \textbf{95.24} & \textbf{78.28} & \textbf{97.59} & \textbf{83.38} & \textbf{95.60} \\
\midrule
{\multirow{2}{*}{{\makecell{Video 2}}}} & SAM3 (person) & 51.09 & 31.24 & 54.97 & 38.48 & 92.36 & 72.79 & 95.21 & 79.33 & 91.33 & 92.26 & \textbf{75.70} & 94.55 & 79.10 & 91.05 \\
& Ours (P-D-DETR) & \textbf{95.29} & \textbf{83.61} & \textbf{96.79} & \textbf{93.03} & \textbf{93.17} & \textbf{73.69} & \textbf{95.99} & \textbf{82.19} & \textbf{93.62} & \textbf{93.06} & 72.16 & \textbf{96.15} & \textbf{83.85} & \textbf{94.04} \\
\bottomrule
\end{tabular}
\end{table*}

\begin{table*}[h]
\scriptsize
\centering
\caption{Ablation study on tracking and multi-view association augmentation on our dataset of real surgeries. We use the off-the-shelf and fine-tuned whole-body detectors (P-D-DETR) as the baseline. The recall of the baseline is an upper bound of the performance because tracking and multi-view association do not create new boxes. \textbf{Best scores}, \underline{second best scores}.}\label{tab:ablation_component}
\begin{tabular}{c|cccc|ccccc|ccccc}
\toprule%
\multirow{2}{*}{Method} & \multicolumn{4}{c|}{Whole body} & \multicolumn{5}{c|}{Face}  & \multicolumn{5}{c}{Eye} \\
\cmidrule{2-15}
& F3 & P & R & $\text{R}_{\mathrm{hard}}$ & F3 & P & R & $\text{R}_{\mathrm{hard}}$ & HoR & F3 & P & R & $\text{R}_{\mathrm{hard}}$ & HoR \\
\midrule
Baseline (off-the-shelf) & \underline{84.42} & 65.25 & \textbf{87.27} & \textbf{66.55} & \textbf{92.35} & 74.80 & \textbf{94.82} & \textbf{73.54} & \textbf{91.14} & \textbf{92.29} & 76.07 & \textbf{94.53} & \textbf{72.87} & \textbf{90.98} \\
+ tracking (forward only) & 79.23 & \textbf{93.17} & 77.93 & 45.02 & 89.88 & \textbf{78.87} & 91.30 & 59.04 & 85.71 & 89.70 & \textbf{79.13} & 91.05 & 57.98 & 85.46 \\
+ tracking (bidirectional) & 80.97 & \underline{91.93} & 79.91 & 49.18 & 90.60 & \underline{78.42} & 92.19 & 62.90 & 87.13 & 90.43 & \underline{78.60} & 91.97 & 62.10 & 86.88 \\
+ multi-view association & \textbf{85.07} & 73.40 & \underline{86.60} & \underline{64.80} & \underline{92.33} & 76.68 & \underline{94.47} & \underline{71.68} & \underline{90.63} & \underline{92.20} & 76.23 & \underline{94.40} & \underline{72.47} & \underline{90.85} \\
\midrule
Baseline (fine-tuned) & \underline{94.63} & 83.76 & \textbf{96.01} & \textbf{89.20} & \underline{94.26} & 75.11 & \textbf{97.01} & \textbf{82.98} & \textbf{94.95} & \underline{94.20} & 76.94 & \textbf{96.61} & \textbf{81.52} & \textbf{94.26} \\
+ tracking (forward only) & 92.10 & \textbf{94.05} & 91.89 & 77.29 & 93.65 & \textbf{78.05} & 95.78 & 77.53 & 93.06 & 93.54 & \underline{77.83} & 95.69 & 77.79 & 92.97 \\
+ tracking (bidirectional) & 93.27 & \underline{93.51} & 93.24 & 81.08 & 93.91 & \underline{77.68} & 96.14 & 79.12 & 93.66 & 93.81 & \textbf{78.64} & 95.87 & 78.32 & 93.19 \\
+ multi-view association & \textbf{94.84} & 85.86 & \underline{95.95} & \underline{88.98} & \textbf{94.30} & 75.55 & \underline{96.97} & \underline{82.85} & \underline{94.92} & \textbf{94.22} & 77.35 & \underline{96.56} & \underline{81.25} & \underline{94.20} \\
\bottomrule
\end{tabular}
\end{table*}

\begin{figure*}[!t]
\centerline{\includegraphics[width=0.9\textwidth]{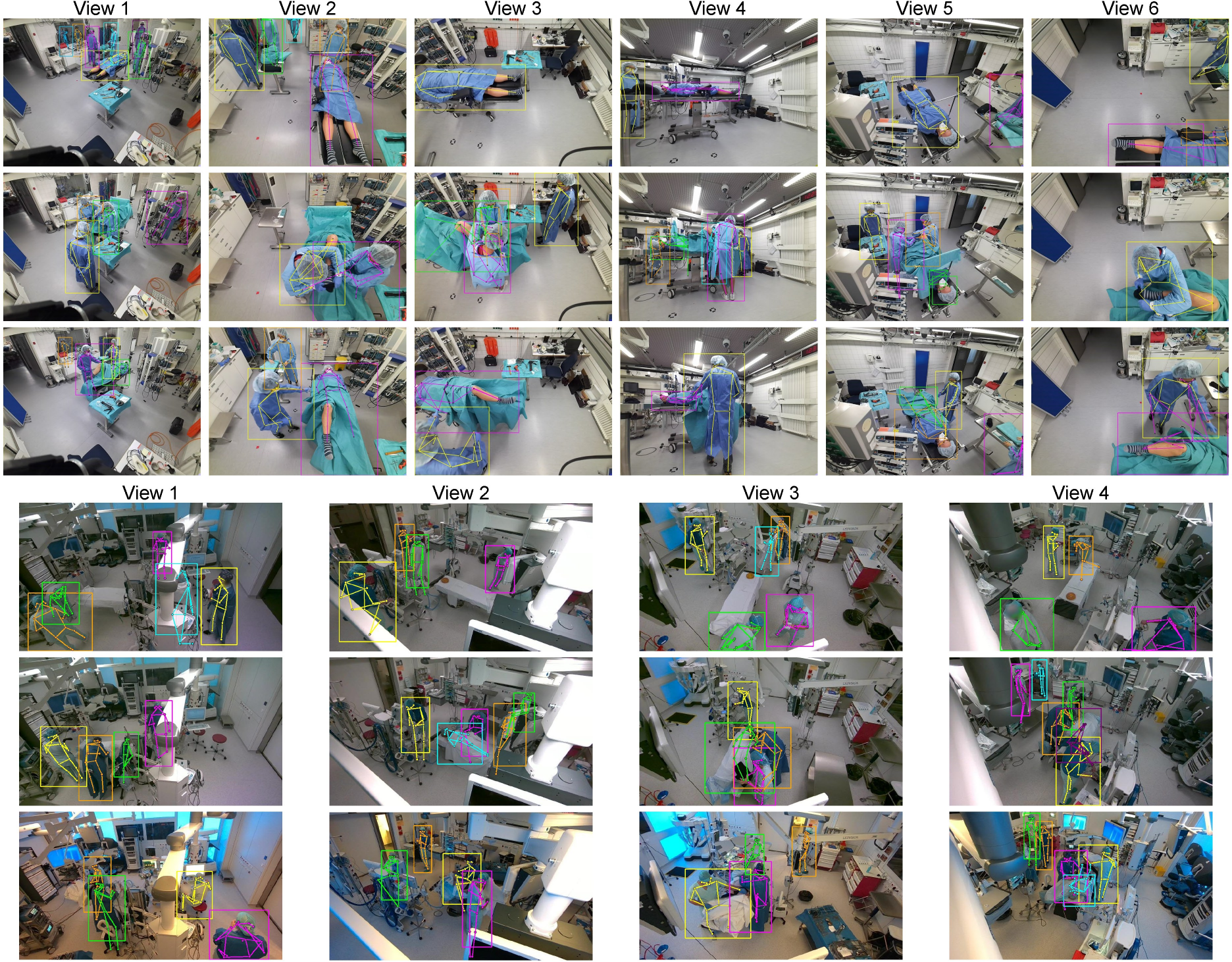}}
\caption{Qualitative results of our approach on 4D-OR (6 views) and our dataset (4 views) with whole-body bounding boxes and 17-joint poses. }
\label{fig:all_results}
\end{figure*}

\subsection{Inference pipeline}

During inference, we apply the fine-tuned whole-body detector on the OR videos to collect a large set of detections with a low-score threshold. Then, we apply tracking in each view to obtain the tracked boxes. 
%In practice, we conduct tracking in two directions: from start to end and from end to start. 
Afterwards, we use tracked boxes in each frame as queries to conduct multi-view association, and obtain the associated boxes. By merging the tracked boxes and associated boxes, we obtain the final whole-body detections. Finally, we apply whole-body pose estimation to localize the keypoints. 

\iffalse
\subsubsection{Manual review}

Although our approach can automatically locate the persons in the OR with high quality, manual review is still needed through the MOSaiC platform~\citep{mazellier2023mosaic} to ensure that all the persons in the OR are properly anonymized with an acceptable number of false positives. As the tracking and multi-view association already assigns unique identity labels for each person, our reviewers can easily remove a batch of false positives using these labels. 
\fi

\subsection{Implementation details}

We use the PyTorch framework to implement our approach with a single NVIDIA RTX 6000 GPU on the Ubuntu system. For tracking, we define detections with scores greater than 0.6 as the high-score set $\mathcal{D}^h$. During tracking and multi-view association, we consider detections with scores greater than 0.1 ($\mathcal{D}^l$) valid for retrieval. We train the geometric encoder for 160 epochs, using the Adam optimizer with an initial learning rate of 1e-4, which decreases to 1e-5 at the 120th epoch. For positional encoding, we set $N$ to 128 in \eqref{eq:pos_encoding1}, and the camera embedding size $V$ is set to 256. Following previous work~\citep{chen2025learning}, the sampling range $[t_{\mathrm{min}},t_{\mathrm{max}}]$ is set to $[5,20]$. The $\alpha$ in \eqref{eq:dis} is set to 0.5. The $\triangle$ in \eqref{eq:triplet_syn} is set to 1.0. For whole-body detection, we fine-tune the detector for 12 epochs in each round, using the AdamW optimizer~\citep{loshchilov2019decoupled} with an initial learning rate of 2e-4, which decreases to 2e-5 at the 10th epoch. For whole-body pose estimation, we fine-tune the model for 90 epochs, using AdamW optimizer with an initial learning rate of 5e-6, which decreases to 5e-7 at the 50th epoch. We train the real-time whole-body detector for 120 epochs using the AdamW optimizer. NMS threshold in \eqref{eq:final_boxes} is set to 0.6. We sample frames at 0.1 FPS to fine-tune the whole-body detector (with the entire CrowdHuman dataset) and the pose detector. 

% \subsubsection{Inference pipeline} During inference, we first use the off-the-shelf whole-body detector to process the whole video. Then, we apply the tracking-assisted detection to find the missing persons in both directions: from start to end and from end to start. Finally, we conduct whole-body pose estimation for all detected persons for further anonymization. 

\section{Experiments}

\subsection{Datasets and evaluation metrics}

We conduct experiments on the 4D-OR~\citep{ozsoy20224d} and our datasets. 4D-OR comprises ten simulated knee surgery videos of 1 Frames Per Second (FPS) recorded with six cameras, resulting in a total of 6,734 frames. Our dataset is composed of five real laparoscopic surgeries of 15 FPS recorded with four cameras, with a total duration of 10.63 hours. For both datasets, we annotate two videos for testing, where each video represents an independent surgery. 

The original 4D-OR dataset does not contain facial keypoints or accurate human bounding box annotations. Therefore, we manually annotate the 4D-OR dataset and our dataset: for each person, we annotate (1) the whole-body bounding box, (2) the ``hard case'' flag (over 67\% occlusion), and (3) three keypoints (eyes and chin) if they are visible. 
The 67\% threshold corresponds to the upper third of occlusion severity when the cases are divided into easy, medium, and hard levels.
In each multi-view image set, we also annotate the identity labels. We annotate the 4D-OR dataset at 0.25 FPS, and annotate our videos at 0.1 FPS. In the end, the 4D-OR test set contains 2,070 images, and ours contains $2,400$ images. 

Two annotators perform the manual annotation. We first use an off-the-shelf person detector~\citep{zheng2022progressive} to initialize the person bounding boxes and a self-supervised multi-view association model~\citep{chen2025learning} to initialize the multi-view identity labels. The annotators then manually correct the bounding boxes and identity labels by comparing the synchronized views, and annotate the facial keypoints and hard-case labels. Before the full annotation, both annotators independently annotate 10 randomly selected multi-view sets following the same annotation rules. Then, they compare their annotations and resolve disagreements to establish consistent annotation criteria. During annotation, they discuss ambiguous cases together and, if necessary, with the project managers. After completing their assigned subsets, each annotator reviews the other's annotations and corrects any remaining inconsistencies.

% We evaluate our approach at three levels: whole-body, face, and eye detection. Since video anonymization does not require precise keypoint localization, we use bounding box-based evaluation metrics for face and eye detection, by manually defining pseudo bounding box labels of a fixed size ($40$x$40$ pixels) centered on the ground-truth face and eye locations. Following prior work~\citep{issenhuth2019face}, we use an IoU threshold of 0.3 to determine correct detections. To ensure fair evaluation, when the ground-truth face or eye regions are not visible in a frame due to occlusion or head orientation, the corresponding boxes are excluded from the computation.

We evaluate our approach at three levels: whole-body, face, and eye detection. 
Since video anonymization does not require precise keypoint localization, we use bounding box-based evaluation metrics for face and eye detection by constructing fixed-size pseudo boxes centered on the facial keypoints. Based on statistic analysis of the annotations, we set the box size to $40$x$40$ pixels as it provides sufficient tolerance to keypoint localization deviations without unnecessarily enlarging the facial region. 
Following prior work~\citep{issenhuth2019face}, we use an IoU threshold of 0.3 to determine correct detections. To ensure fair evaluation, when the ground-truth face or eye regions are not visible in a frame due to occlusion or head orientation, the corresponding boxes are excluded from the computation.

We evaluate our approach using five metrics: F3-score (F3), precision (P), recall (R), hard-case recall ($\text{R}_{\mathrm{hard}}$), and holistic recall (HoR)~\citep{bastian2023disguisor}. F3 is the $\text{F}_\beta$ score with $\beta=3$, which gives substantially greater weight to R than P. In the context of video anonymization, R is more important than P in order not to miss a person. Therefore, we automatically select the score threshold to maximize F3 for each method, and report corresponding P, R, $\text{R}_{\mathrm{hard}}$, and HoR. $\text{R}_{\mathrm{hard}}$ explicitly evaluates models' capability to detect heavily occluded cases, while HoR further showcases the unique challenges of multi-camera setups, where a subject is only considered correctly anonymized if they are detected in every view in which they are visible within a multi-view set.

\begin{table*}[h]
\scriptsize
\centering
\caption{Ablation study on fine-tuning the P-D-DETR model iteratively on our dataset of real surgeries.}\label{tab:ablation_iter}
\begin{tabular}{c|cccc|ccccc|ccccc}
\toprule%
\multirow{2}{*}{Method} & \multicolumn{4}{c|}{Whole body} & \multicolumn{5}{c|}{Face}  & \multicolumn{5}{c}{Eye} \\
\cmidrule{2-15}
& F3 & P & R & $\text{R}_{\mathrm{hard}}$ & F3 & P & R & $\text{R}_{\mathrm{hard}}$ & HoR & F3 & P & R & $\text{R}_{\mathrm{hard}}$ & HoR \\
\midrule
Baseline & 84.42 & 65.25 & 87.27 & 66.55 & 92.35 & 74.80 & 94.82 & 73.54 & 91.14 & 92.29 & 76.07 & 94.53 & 72.87 & 90.98 \\
1st iteration & 91.28 & 82.83 & 92.32 & 77.55 & 93.99 & \textbf{77.16} & 96.32 & 79.12 & 93.79 & 93.96 & \textbf{77.91} & 96.16 & 78.99 & 93.63 \\
2nd iteration  & \textbf{94.63} & \textbf{83.76} & \textbf{96.01} & \textbf{89.20} & \textbf{94.26} & 75.11 & \textbf{97.01} & \textbf{82.98} & \textbf{94.95} & \textbf{94.20} & 76.94 & \textbf{96.61} & \textbf{81.52} & \textbf{94.26} \\
3rd iteration  & 93.52 & 83.07 & 94.85 & 86.25 & 93.76 & 76.81 & 96.12 & 79.26 & 93.34 & 93.62 & 75.77 & 96.14 & 80.45 & 93.38 \\
\bottomrule
\end{tabular}
\end{table*}

\begin{table*}[h]
\scriptsize
\centering
\caption{Ablation study on low-score, high-score, and NMS thresholds on our dataset of real surgeries.}\label{tab:ablation_threshold}
\begin{tabular}{c|cccc|ccccc|ccccc}
\toprule%
\multirow{2}{*}{[Low, high, NMS]} & \multicolumn{4}{c|}{Whole body} & \multicolumn{5}{c|}{Face}  & \multicolumn{5}{c}{Eye} \\
\cmidrule{2-15}
& F3 & P & R & $\text{R}_{\mathrm{hard}}$ & F3 & P & R & $\text{R}_{\mathrm{hard}}$ & HoR & F3 & P & R & $\text{R}_{\mathrm{hard}}$ & HoR \\
\midrule
$[$0.1, 0.6, 0.6$]$ (baseline) & \textbf{94.84} & 85.86 & \textbf{95.95} & \textbf{88.98} & \textbf{94.30} & 75.55 & \textbf{96.97} & 82.85 & \textbf{94.92} & \textbf{94.22} & 77.35 & \textbf{96.56} & 81.25 & \textbf{94.20} \\
\midrule
$[$\textbf{0.2}, 0.6, 0.6$]$ & 92.65 & 89.66 & 93.00 & 79.70 & 93.97 & 77.24 & 96.28 & 79.39 & 93.72 & 93.88 & \textbf{78.11} & 96.03 & 79.12 & 93.34 \\
$[$\textbf{0.3}, 0.6, 0.6$]$ & 91.72 & \textbf{91.64} & 91.73 & 76.11 & 93.85 & \textbf{78.43} & 95.94 & 76.46 & 93.19 & 93.83 & 77.98 & 95.99 & 77.53 & 93.31 \\
\midrule
$[$0.1, \textbf{0.5}, 0.6$]$  & 94.32 & 85.74 & 95.38 & 87.48 & 94.00 & 74.30 & 96.85 & \textbf{84.44} & 94.57 & 93.97 & 76.00 & 96.50 & \textbf{83.64} & 94.04 \\
$[$0.1, \textbf{0.7}, 0.6$]$  & 94.32 & 85.61 & 95.40 & 87.25 & 94.03 & 76.74 & 96.45 & 80.85 & 93.79 & 93.97 & 77.37 & 96.27 & 80.98 & 93.50 \\
\midrule
$[$0.1, 0.6, \textbf{0.4}$]$  & 93.44 & 83.26 & 94.73 & 85.24 & 93.84 & 76.46 & 96.27 & 79.79 & 93.69 & 93.98 & 76.58 & 96.41 & 81.12 & 93.97 \\
$[$0.1, 0.6, \textbf{0.8}$]$  & 94.10 & 82.61 & 95.57 & 88.06 & 94.08 & 76.26 & 96.59 & 81.12 & 94.26 & 94.11 & 76.97 & 96.50 & 81.25 & 94.10 \\
\bottomrule
\end{tabular}
\end{table*}

\subsection{Results}

We compare our method with three categories of anonymization approaches: face detection, head detection, and person detection followed by pose estimation. Specifically, we compare our method with: (1) RetinaFace model~\citep{deng2020retinaface} trained on the WIDER FACE dataset~\citep{yang2016wider} for face detection; (2) YOLOv13 model~\citep{lei2025yolov13} trained on the CrowdHuman dataset~\citep{shao2018crowdhuman} for head detection; (3) Progressive Deformable DETR (P-D-DETR) model~\citep{zheng2022progressive} trained on the CrowdHuman dataset~\citep{shao2018crowdhuman} for whole-body detection; (4) SAM3 model~\citep{carion2026sam} with text prompts of ``face'', ``head'', and ``person'', respectively; (5) self-supervised Mean Teacher~\citep{hao2024simplifying}; (6) self-supervised Iter-Score~\citep{issenhuth2019face}. SAM3 serves as a particularly strong zero-shot baseline because it is a recent promptable vision foundation model pretrained on large-scale datasets. P-D-DETR, SAM3 (person), and the self-supervised approaches all follow the same two-stage pipeline as ours, consisting of whole-body detection followed by pose estimation. For a fair comparison, all these two-stage approaches use the same fine-tuned pose detector. For our approach and Iter-Score, we train the models using pseudo-labels initialized from P-D-DETR or SAM3 predictions. We also train a real-time detector, DEIM~\citep{huang2025deim}, to demonstrate its real-time applicability. For fair comparison, the evaluation strategy of face and eye detection is applied to the outputs of all methods.

% We first compare our method with two different types of anonymization approaches: face detection and head detection. For face detection, we compare with the state-of-the-art RetinaFace model~\citep{deng2020retinaface} trained on the WIDER FACE dataset~\citep{yang2016wider}. For head detection, we compare with a YOLOv8 model~\citep{varghese2024yolov8} trained on the CrowdHuman dataset~\citep{shao2018crowdhuman}. Furthermore, for whole-body detection, we use the Progressive Deformable DETR (P-D-DETR) model~\citep{zheng2022progressive} trained on the Crowdhuman dataset as the baseline, and compare our approach with a self-supervised detection approach in~\citep{issenhuth2019face}, where we call it Iter-Score. To test the real-time applicability, we train a state-of-the-art real-time detector, DEIM~\citep{huang2025deim}, using our generated pseudo-labels. For fair comparison, the evaluation strategy of face and eye detection is applied to the outputs of all methods. 

Table~\ref{tab:results} presents the quantitative results on the 4D-OR and our datasets.
On 4D-OR, both our approach and SAM3 achieve approximately 99\% recall for face and eye detection. However, on our dataset of real surgeries, the performance of all the approaches drops in different degrees. In particular, SAM3 (face) drops from 99\% to less than 93\%. This further indicates the importance of evaluating anonymization methods on real surgical data rather than relying exclusively on simulated benchmarks.
Our method achieves approximately 97\% recall, outperforming the other approaches, which demonstrates the necessity of domain adaptation in real OR scenes. Although Iter-Score~\citep{issenhuth2019face} is capable of achieving the best precision, it fails to detect hard cases with a low recall rate. 
It is also noticeable that compared to direct face or head detection approaches, whole body detection followed by pose detection achieves significantly better performance on the real OR videos, showing the advantage of this two-stage pipeline for addressing challenging cases. Table~\ref{tab:single_video_test} further reports video-level test results of SAM3 and our approach on our dataset. While SAM3 exhibits a large performance variation between these two videos, our method shows more consistent performance. Qualitative examples of different anonymization approaches are shown in Fig.~\ref{fig:results}. We also display more qualitative results of our approach on the 4D-OR dataset and our dataset in Fig.~\ref{fig:all_results}. 

To evaluate real-time applicability, we train DEIM using our framework, which achieves comparable performance at 52.09 FPS. During full inference including person detection, tracking, multi-view association, and pose estimation, the DEIM-based pipeline achieves 22.85 FPS. 
Compared to SAM3 (face), our approach improves recall by over 4\% with better precision. Since each video in our dataset is recorded at 15 FPS and contains approximately five persons per frame, our approach avoids at least 10,000 missed detections in a one-hour video. Assuming it takes 4 seconds to detect and blur a missed person, which is an optimistic estimate based on a small-scale timing test, our approach conservatively saves 10 hours of manual correction time.

\begin{table}[t]
\scriptsize
\centering
\caption{Ablation study on fine-tuning (FT) the whole-body human pose detector using generated pseudo keypoint labels on our dataset of real surgeries.}\label{tab:abl_kpt}
\begin{tabular}{c|cccc|cccc}
\toprule%
\multirow{2}{*}{FT} & \multicolumn{4}{c|}{Face}  & \multicolumn{4}{c}{Eye} \\
\cmidrule{2-9}
 & P & R & $\text{R}_{\mathrm{hard}}$ & HoR & P & R & $\text{R}_{\mathrm{hard}}$ & HoR \\
\midrule
\xmark & 75.09 & 94.45 & 75.53 & 90.91 & 75.63 & 94.49 & 75.80 & 90.91 \\
\checkmark & \textbf{75.55} & \textbf{96.97} & \textbf{82.85} & \textbf{94.92} & \textbf{77.35} & \textbf{96.56} & \textbf{81.25} & \textbf{94.20} \\
\bottomrule
\end{tabular}
\end{table}

\subsection{Ablation study}
\subsubsection{Tracking and multi-view person association}
To evaluate the effectiveness of tracking and multi-view association, we conduct an ablation study on our dataset. We set up two sets of experiments using the off-the-shelf and fine-tuned whole-body detectors as the baselines to generate the candidate pool. Since tracking and multi-view association do not create new boxes, the recall of the baselines represents the theoretical upper bound of the performance, because the baselines use all the boxes in $\mathcal{D}^h$ and $\mathcal{D}^l$. Results are shown in Table~\ref{tab:ablation_component}. For the off-the-shelf detector, applying tracking significantly boosts precision, but degrades recall dramatically. After adding multi-view association, we achieve a better recall rate close to the upper bound, with much better precision than the baseline. By adopting such a trade-off strategy, our method retrieves the maximum number of true positives with a manageable false positive rate, and thus the final boxes can be used as pseudo-labels to fine-tune the detector to retrieve more false negatives. In comparison, fine-tuning only with high-score boxes~\citep{issenhuth2019face} yields much lower recall, as shown in Table~\ref{tab:results}.

%Moreover, it is noticeable that for the iteratively fine-tuned detector, applying tracking and multi-view person association does not make very obvious changes anymore for face and eye detection. Therefore, after the detector is well fine-tuned, the inference pipeline can be further simplified by removing the tracking and multi-view person association modules. 

\subsubsection{Iterative whole-body detector fine-tuning}
We conduct an ablation study on the iterative training strategy on our dataset. As shown in Table~\ref{tab:ablation_iter}, after the first iteration, the detector performs much better with higher precision and recall. After the second iteration, the detector achieves the best performance. After the third iteration, there is a slight performance degradation, which is mainly caused by the accumulation of noise in the pseudo-labels. Therefore, in our dataset of real surgeries, we find that fine-tuning the detector with two iterative rounds is enough for optimal performance.

% We conduct an ablation study on the iterative training strategy on our dataset. As shown in Table~\ref{tab:ablation_iter}, after the first iteration, the detector performs much better with higher precision and recall. After the second iteration, the detector achieves the best performance. After the third iteration, although the detector obtains the highest AP on face and eye detection, the more important recall rate does not get boosted anymore. Therefore, in our dataset of real surgeries, we find that fine-tuning the detector with two iterative rounds is enough for optimal performance.

\subsubsection{Low-score, high-score, and NMS thresholds}
We conduct a sensitivity analysis on the low-score, high-score, and NMS thresholds used in pseudo-label generation. As shown in Table~\ref{tab:ablation_threshold}, increasing the low-score threshold generally improves the precision but degrades the recall. The performance is less sensitive to the high-score and NMS thresholds than to the low-score threshold.

\subsubsection{Fine-tuning pose detector}
To evaluate the effectiveness of fine-tuning whole-body pose detector, we compare the fine-tuned model and the off-the-shelf model by applying them on the final whole-body detections. As shown in Table~\ref{tab:abl_kpt}, fine-tuning the model significantly improves the performance. 

\begin{figure}[t]
\centerline{\includegraphics[width=\columnwidth]{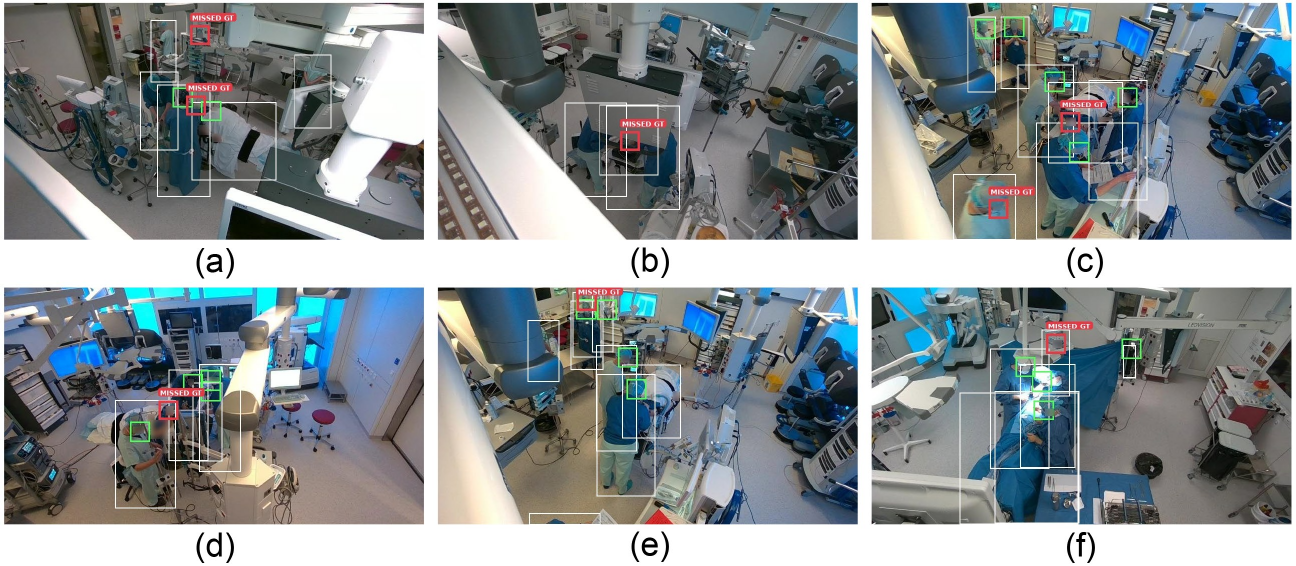}}
\caption{Representative failure cases, where white boxes denote whole-body detections, green boxes denote pseudo face boxes generated from detected facial keypoints, and red boxes denote missed ground-truth face boxes. There are five failure patterns: (1) missed whole-body detection in (a); (2) missed keypoint detection in (b), (c), and (f); (3) duplicated keypoint detection in (c) and (e); (4) merged whole-body detection in (d); and (5) inaccurate keypoint detection in (a).}
\label{fig:failure}
\end{figure}

\subsection{Failure analysis}

We analyze the remaining 3\% missed detections on the real surgical videos and present representative qualitative examples in Fig.~\ref{fig:failure}. We identify five failure patterns: (1) missed whole-body detection, where the person is not detected; (2) missed keypoint detection, where a person bounding box is available but the corresponding facial keypoints are not detected; (3) duplicated keypoint detection, where two nearby persons are correctly detected but the pose detector localizes both sets of facial keypoints on the same person; (4) merged whole-body detection, where two nearby persons are represented by a single bounding box; (5) inaccurate keypoint detection, where the predicted facial keypoints are spatially inaccurate.

To further disentangle errors introduced by whole-body detection and pose estimation, we conduct an oracle experiment in which the detected person boxes are associated with the ground-truth facial keypoints they cover using Hungarian matching. Under this oracle setting, 99\% of the ground-truth facial keypoints are successfully associated with a detected person box. This shows that for the remaining 3\% missed detections, 1\% comes from person detection and 2\% comes from pose detection.

\subsection{Discussion and limitations}

In this work, we investigate multi-view video anonymization using real surgical videos involving human patients. To the best of our knowledge, we provide the first evaluation of several video anonymization approaches under such conditions. The substantial performance gap between simulated and real surgical videos highlights the importance of evaluation in real clinical environments. Our test set contains $2,400$ images, which is comparable in scale to established benchmarks in this domain~\citep{flouty2018faceoff, issenhuth2019face}. However, it consists of only two laparoscopic surgery videos collected at a single clinical site, which limits the diversity of scene layouts, surgical procedures, and OR team size. Due to the strict ethical regulations governing the collection of real surgical data, it is very challenging to expand the dataset, which is also the main motivation for studying video anonymization. By taking this initial step, we hope to inspire further related research and facilitate the future collection of more diverse real surgical datasets.

Although our approach achieves strong performance on real surgical videos, there are some limitations. 
First, the multi-view association strategy requires both synchronized multi-view data and stationary cameras. Although multi-view camera setups are increasingly common in modern ORs~\citep{belagiannis2016parsing,srivastav2018mvor,ozsoy20224d,bastian2023disguisor,ozsoy2025mm}, older ORs may be limited to a single camera. We observe that introducing a $1.0$-second temporal offset in the training data still yields approximately 90\% association accuracy. In contrast, a camera rotation of approximately $30^\circ$ reduces the association accuracy from around 90\% to 50\%. Therefore, if the camera placement changes, the association model must be retrained. If the cameras are continuously moving in a video, the multi-view association strategy would not be applicable. One possible solution for moving-camera settings is to model and update camera poses over time. Nevertheless, when only a single camera is available, or camera poses change continuously, our pipeline remains deployable by bypassing the multi-view association module and relying on single-view tracking.
Second, after the second iteration of training, the model no longer improves, while there are still approximately 3\% missed detections. This is mainly because of the accumulation of noise in the pseudo-labels, which is a common limitation of iterative self-training. Several directions could be explored to address this issue: (1) 3D geometric priors could be introduced to further estimate the uncertainty of the pseudo-labels; (2) person detection and pose estimation could be integrated into an end-to-end learning framework; (3) the emerging vision foundation models could be further incorporated to exploit their strong zero-shot generalization capabilities for pseudo-label generation.
Lastly, when deploying our approach at a new clinical site without labeled validation data, the optimal hyperparameter settings cannot be determined directly. However, our sensitivity analyses in Tables~\ref{tab:ablation_iter} and~\ref{tab:ablation_threshold} show that the performance is relatively insensitive to the high-score and NMS thresholds, while the low-score threshold should remain low. We also observe comparable performance with one or two self-training iterations. For multi-view association, triplet margins from 0.5 to 2.0 all result in around 90\% association accuracy, showing limited sensitivity to this parameter. Therefore, we recommend using the same fixed settings as in this work when labeled validation data are unavailable.

% Although our approach achieves remarkable performance in real surgical videos, there are some limitations: (1) if a person is severely occluded in every view for a period of time, or if it only appears for a few frames, our approach could fail to detect it, making the remaining 3\% of persons undetected; (2) multi-view geometry learning requires synchronized camera views with fixed positions; (3) ``ghost association'' could appear if a person looks too similar with the background. We envision that these constraints could be overcome with more advanced tracking and multi-view geometry techniques in the future. 

\section{Conclusion}

We present a self-supervised approach to address the challenging problem of video anonymization in the operating room (OR). Traditional face detection in the OR suffers from high false negative rates due to severe occlusions from medical equipment and obstruction of facial features by clinicians wearing masks and caps. Therefore, we design a two-stage pipeline that includes whole-body detection and whole-body pose estimation to take advantage of more redundant features against the dramatic domain gap. During the first stage, we utilize tracking and multi-view association to retrieve the missed detections using temporal and multi-view context, and then fine-tune the whole-body detector iteratively using generated pseudo-labels. During the second stage, we fine-tune the whole-body pose detector using its own predictions. 
Experimental results demonstrate that our approach achieves 99\% recall on the simulated 4D-OR dataset and 97\% recall on our real-surgery dataset. The improved recall on real surgical videos corresponds to a conservative saving of 10 hours of manual correction time for a one-hour video. 
We hope that this work inspires further research that facilitates the collection of more real surgical video recordings across diverse clinical settings, ultimately advancing the development of automated, context-aware support systems in the modern ORs.

\section{Acknowledgments}

This work was supported by French state funds managed within the Plan Investissements d’Avenir by the ANR under references ANR-22-FAI1-0001 (project DAIOR), ANR-10-IAHU-02 (IHU Strasbourg) and by BPI France (Project 5G-OR).
This work was also granted access to the servers/HPC resources managed by CAMMA, IHU Strasbourg, Unistra Mesocentre, and GENCI-IDRIS [AD011014722R2, AD011011631R4, and AD011011638R3].

\bibliographystyle{model2-names.bst}
\bibliography{tmi}

@article{padoy2019machine,
  title={Machine and deep learning for workflow recognition during surgery},
  author={Padoy, Nicolas},
  journal={Minimally Invasive Therapy \& Allied Technologies},
  volume={28},
  number={2},
  pages={82--90},
  year={2019},
  publisher={Taylor \& Francis},
  doi = {10.1080/13645706.2019.1584116}
}

@article{mascagni2021or,
  title={OR black box and surgical control tower: recording and streaming data and analytics to improve surgical care},
  author={Mascagni, Pietro and Padoy, Nicolas},
  journal={Journal of Visceral Surgery},
  volume={158},
  number={3},
  pages={S18--S25},
  year={2021},
  publisher={Elsevier},
  doi = {10.1016/j.jviscsurg.2021.01.004}
}

@article{srivastav2022unsupervised,
  title={Unsupervised domain adaptation for clinician pose estimation and instance segmentation in the operating room},
  author={Srivastav, Vinkle and Gangi, Afshin and Padoy, Nicolas},
  journal={Medical image analysis},
  volume={80},
  pages={102525},
  year={2022},
  publisher={Elsevier},
  doi={10.1016/j.media.2022.102525}
}

@InProceedings{srivastav2018mvor,
  title      = "MVOR: A Multi-view RGB-D Operating Room Dataset for 2D and 3D Human Pose Estimation",
  author     = "Srivastav, Vinkle and Issenhuth, Thibaut and Kadkhodamohammadi Abdolrahim and de Mathelin, Michel and Gangi, Afshin and Padoy, Nicolas",
  conference = "MICCAI-LABELS Workshop",
  year       = "2018",
  doi = {10.48550/arXiv.1808.08180},
}

@inproceedings{ozsoy20224d,
  title={4d-or: Semantic scene graphs for or domain modeling},
  author={{\"O}zsoy, Ege and {\"O}rnek, Evin P{\i}nar and Eck, Ulrich and Czempiel, Tobias and Tombari, Federico and Navab, Nassir},
  booktitle={International Conference on Medical Image Computing and Computer-Assisted Intervention},
  pages={475--485},
  year={2022},
  organization={Springer}, 
  doi = {10.1007/978-3-031-16449-1_45}
}

@article{belagiannis2016parsing,
  title={Parsing human skeletons in an operating room},
  author={Belagiannis, Vasileios and Wang, Xinchao and Shitrit, Horesh Beny Ben and Hashimoto, Kiyoshi and Stauder, Ralf and Aoki, Yoshimitsu and Kranzfelder, Michael and Schneider, Armin and Fua, Pascal and Ilic, Slobodan and others},
  journal={Machine Vision and Applications},
  volume={27},
  pages={1035--1046},
  year={2016},
  publisher={Springer},
  doi = {10.1007/s00138-016-0792-4}
}

@article{mazellier2023mosaic,
  title={MOSaiC: a Web-based Platform for Collaborative Medical Video Assessment and Annotation},
  author={Mazellier, Jean-Paul and Boujon, Antoine and Bour-Lang, M{\'e}line and Erharhd, Ma{\"e}l and Waechter, Julien and Wernert, Emilie and Mascagni, Pietro and Padoy, Nicolas},
  journal={arXiv},
  year={2023},
  doi = {10.48550/arXiv.2312.08593}
}

@inproceedings{zhang2022bytetrack,
  title={Bytetrack: Multi-object tracking by associating every detection box},
  author={Zhang, Yifu and Sun, Peize and Jiang, Yi and Yu, Dongdong and Weng, Fucheng and Yuan, Zehuan and Luo, Ping and Liu, Wenyu and Wang, Xinggang},
  booktitle={European conference on computer vision},
  pages={1--21},
  year={2022},
  organization={Springer},
  doi = {10.1007/978-3-031-20047-2_1}
}

@inproceedings{czempiel2022surgical,
  title={Surgical workflow recognition: from analysis of challenges to architectural study},
  author={Czempiel, Tobias and Sharghi, Aidean and Paschali, Magdalini and Navab, Nassir and Mohareri, Omid},
  booktitle={European Conference on Computer Vision},
  pages={556--568},
  year={2022},
  organization={Springer},
  doi={10.1007/978-3-031-25066-8_32}
}

@article{li2020robotic,
  title={A robotic 3d perception system for operating room environment awareness},
  author={Li, Zhaoshuo and Shaban, Amirreza and Simard, Jean-Gabriel and Rabindran, Dinesh and DiMaio, Simon and Mohareri, Omid},
  journal={arXiv preprint arXiv:2003.09487},
  year={2020},
  doi={10.48550/arXiv.2003.09487}
}

@article{hu2022multi,
  title={Multi-camera multi-person tracking and re-identification in an operating room},
  author={Hu, Haowen and Hachiuma, Ryo and Saito, Hideo and Takatsume, Yoshifumi and Kajita, Hiroki},
  journal={Journal of Imaging},
  volume={8},
  number={8},
  pages={219},
  year={2022},
  publisher={MDPI},
  doi={10.3390/jimaging8080219}
}

@inproceedings{ozsoy2023labrad,
  title={Labrad-or: lightweight memory scene graphs for accurate bimodal reasoning in dynamic operating rooms},
  author={{\"O}zsoy, Ege and Czempiel, Tobias and Holm, Felix and Pellegrini, Chantal and Navab, Nassir},
  booktitle={International Conference on Medical Image Computing and Computer-Assisted Intervention},
  pages={302--311},
  year={2023},
  organization={Springer},
  doi={10.1007/978-3-031-43996-4_29}
}

@article{ozsoy2024holistic,
  title={Holistic OR domain modeling: a semantic scene graph approach},
  author={{\"O}zsoy, Ege and Czempiel, Tobias and {\"O}rnek, Evin P{\i}nar and Eck, Ulrich and Tombari, Federico and Navab, Nassir},
  journal={International Journal of Computer Assisted Radiology and Surgery},
  volume={19},
  number={5},
  pages={791--799},
  year={2024},
  publisher={Springer},
  doi={10.1007/s11548-023-03022-w}
}

@article{pei2024s2former,
  title={S$^2$Former-OR: Single-Stage Bi-Modal Transformer for Scene Graph Generation in OR},
  author={Pei, Jialun and Guo, Diandian and Zhang, Jingyang and Lin, Manxi and Jin, Yueming and Heng, Pheng-Ann},
  journal={IEEE Transactions on Medical Imaging},
  volume={44},
  number={1},
  pages={361--372},
  year={2024},
  publisher={IEEE},
  doi={10.1109/TMI.2024.3444279}
}

@article{hansen2019fusing,
  title={Fusing information from multiple 2D depth cameras for 3D human pose estimation in the operating room},
  author={Hansen, Lasse and Siebert, Marlin and Diesel, Jasper and Heinrich, Mattias P},
  journal={International journal of computer assisted radiology and surgery},
  volume={14},
  pages={1871--1879},
  year={2019},
  publisher={Springer},
  doi={10.1007/s11548-019-02044-7}
}

@inproceedings{flouty2018faceoff,
  title={FaceOff: Anonymizing Videos in the Operating Rooms},
  author={Flouty, Evangello and Zisimopoulos, Odysseas and Stoyanov, Danail},
  booktitle={OR 2.0/CARE/CLIP/ISIC@ MICCAI},
  year={2018},
  doi={10.1007/978-3-030-01201-4_4}
}

@article{issenhuth2019face,
  title={Face detection in the operating room: Comparison of state-of-the-art methods and a self-supervised approach},
  author={Issenhuth, Thibaut and Srivastav, Vinkle and Gangi, Afshin and Padoy, Nicolas},
  journal={International journal of computer assisted radiology and surgery},
  volume={14},
  pages={1049--1058},
  year={2019},
  publisher={Springer},
  doi={10.1007/s11548-019-01944-y}
}

@article{bastian2023disguisor,
  title={DisguisOR: holistic face anonymization for the operating room},
  author={Bastian, Lennart and Wang, Tony Danjun and Czempiel, Tobias and Busam, Benjamin and Navab, Nassir},
  journal={International Journal of Computer Assisted Radiology and Surgery},
  volume={18},
  number={7},
  pages={1209--1215},
  year={2023},
  publisher={Springer},
  doi={10.1007/s11548-023-02939-6}
}

@inproceedings{zheng2022progressive,
  title={Progressive end-to-end object detection in crowded scenes},
  author={Zheng, Anlin and Zhang, Yuang and Zhang, Xiangyu and Qi, Xiaojuan and Sun, Jian},
  booktitle={Proceedings of the IEEE/CVF conference on computer vision and pattern recognition},
  pages={857--866},
  year={2022},
  doi={10.1109/CVPR52688.2022.00093}
}

@article{shao2018crowdhuman,
  title={Crowdhuman: A benchmark for detecting human in a crowd},
  author={Shao, Shuai and Zhao, Zijian and Li, Boxun and Xiao, Tete and Yu, Gang and Zhang, Xiangyu and Sun, Jian},
  journal={arXiv preprint arXiv:1805.00123},
  year={2018},
  doi={10.48550/arXiv.1805.00123}
}

@article{tancik2020fourier,
  title={Fourier features let networks learn high frequency functions in low dimensional domains},
  author={Tancik, Matthew and Srinivasan, Pratul and Mildenhall, Ben and Fridovich-Keil, Sara and Raghavan, Nithin and Singhal, Utkarsh and Ramamoorthi, Ravi and Barron, Jonathan and Ng, Ren},
  journal={Advances in neural information processing systems},
  volume={33},
  pages={7537--7547},
  year={2020}
}

@inproceedings{jin2020whole,
  title={Whole-body human pose estimation in the wild},
  author={Jin, Sheng and Xu, Lumin and Xu, Jin and Wang, Can and Liu, Wentao and Qian, Chen and Ouyang, Wanli and Luo, Ping},
  booktitle={European Conference on Computer Vision},
  pages={196--214},
  year={2020},
  organization={Springer},
  doi={10.1007/978-3-030-58545-7_12}
}

@inproceedings{deng2020retinaface,
  title={Retinaface: Single-shot multi-level face localisation in the wild},
  author={Deng, Jiankang and Guo, Jia and Ververas, Evangelos and Kotsia, Irene and Zafeiriou, Stefanos},
  booktitle={Proceedings of the IEEE/CVF conference on computer vision and pattern recognition},
  pages={5203--5212},
  year={2020},
  doi={10.1109/CVPR42600.2020.00525}
}

@inproceedings{yang2016wider,
  title={Wider face: A face detection benchmark},
  author={Yang, Shuo and Luo, Ping and Loy, Chen-Change and Tang, Xiaoou},
  booktitle={Proceedings of the IEEE conference on computer vision and pattern recognition},
  pages={5525--5533},
  year={2016},
  doi={10.1109/CVPR.2016.596}
}

@article{vercauteren2019cai4cai,
  title={Cai4cai: the rise of contextual artificial intelligence in computer-assisted interventions},
  author={Vercauteren, Tom and Unberath, Mathias and Padoy, Nicolas and Navab, Nassir},
  journal={Proceedings of the IEEE},
  volume={108},
  number={1},
  pages={198--214},
  year={2019},
  publisher={IEEE},
  doi={10.1109/JPROC.2019.2946993}
}

@article{maier2022surgical,
  title={Surgical data science--from concepts toward clinical translation},
  author={Maier-Hein, Lena and Eisenmann, Matthias and Sarikaya, Duygu and M{\"a}rz, Keno and Collins, Toby and Malpani, Anand and Fallert, Johannes and Feussner, Hubertus and Giannarou, Stamatia and Mascagni, Pietro and others},
  journal={Medical image analysis},
  volume={76},
  pages={102306},
  year={2022},
  publisher={Elsevier},
  doi={10.1016/j.media.2021.102306}
}

@inproceedings{ozsoy2025mm,
  title={Mm-or: A large multimodal operating room dataset for semantic understanding of high-intensity surgical environments},
  author={{\"O}zsoy, Ege and Pellegrini, Chantal and Czempiel, Tobias and Tristram, Felix and Yuan, Kun and Bani-Harouni, David and Eck, Ulrich and Busam, Benjamin and Keicher, Matthias and Navab, Nassir},
  booktitle={Proceedings of the Computer Vision and Pattern Recognition Conference},
  pages={19378--19389},
  year={2025},
  doi={10.1109/CVPR52734.2025.01805}
}

@inproceedings{chen2025learning,
  title={Learning from Synchronization: Self-Supervised Uncalibrated Multi-View Person Association in Challenging Scenes},
  author={Chen, Keqi and Srivastav, Vinkle and Mutter, Didier and Padoy, Nicolas},
  booktitle={Proceedings of the Computer Vision and Pattern Recognition Conference},
  pages={24419--24428},
  year={2025},
  doi={10.1109/CVPR52734.2025.02274}
}

@article{nwoye2023cholectriplet2022,
  title={Cholectriplet2022: Show me a tool and tell me the triplet—an endoscopic vision challenge for surgical action triplet detection},
  author={Nwoye, Chinedu Innocent and Yu, Tong and Sharma, Saurav and Murali, Aditya and Alapatt, Deepak and Vardazaryan, Armine and Yuan, Kun and Hajek, Jonas and Reiter, Wolfgang and Yamlahi, Amine and others},
  journal={Medical Image Analysis},
  volume={89},
  pages={102888},
  year={2023},
  publisher={Elsevier},
  doi={10.1016/j.media.2023.102888}
}

@article{murali2023endoscapes,
  title={The endoscapes dataset for surgical scene segmentation, object detection, and critical view of safety assessment: Official splits and benchmark},
  author={Murali, Aditya and Alapatt, Deepak and Mascagni, Pietro and Vardazaryan, Armine and Garcia, Alain and Okamoto, Nariaki and Costamagna, Guido and Mutter, Didier and Marescaux, Jacques and Dallemagne, Bernard and others},
  journal={arXiv preprint arXiv:2312.12429},
  year={2023},
  doi={10.48550/arXiv.2312.12429}
}

@article{lavanchy2024challenges,
  title={Challenges in multi-centric generalization: phase and step recognition in Roux-en-Y gastric bypass surgery},
  author={Lavanchy, Jo{\"e}l L and Ramesh, Sanat and Dall’Alba, Diego and Gonzalez, Cristians and Fiorini, Paolo and M{\"u}ller-Stich, Beat P and Nett, Philipp C and Marescaux, Jacques and Mutter, Didier and Padoy, Nicolas},
  journal={International journal of computer assisted radiology and surgery},
  volume={19},
  number={11},
  pages={2249--2257},
  year={2024},
  publisher={Springer},
  doi={10.1007/s11548-024-03166-3}
}

@inproceedings{che2026lemon,
  title={Lemon: A large endoscopic monocular dataset and foundation model for perception in surgical settings},
  author={Che, Chengan and Wang, Chao and Vercauteren, Tom and Tsoka, Sophia and Garcia-Peraza-Herrera, Luis C},
  booktitle={Proceedings of the IEEE/CVF Conference on Computer Vision and Pattern Recognition},
  pages={42659--42669},
  year={2026}
}

@inproceedings{gafni2019live,
  title={Live face de-identification in video},
  author={Gafni, Oran and Wolf, Lior and Taigman, Yaniv},
  booktitle={Proceedings of the IEEE/CVF International Conference on Computer Vision},
  pages={9378--9387},
  year={2019},
  doi={10.1109/ICCV.2019.00947}
}

@inproceedings{maximov2020ciagan,
  title={Ciagan: Conditional identity anonymization generative adversarial networks},
  author={Maximov, Maxim and Elezi, Ismail and Leal-Taix{\'e}, Laura},
  booktitle={Proceedings of the IEEE/CVF conference on computer vision and pattern recognition},
  pages={5447--5456},
  year={2020}, 
  doi={10.1109/CVPR42600.2020.00549}
}

@inproceedings{rosberg2023fiva,
  title={FIVA: facial image and video anonymization and anonymization defense},
  author={Rosberg, Felix and Aksoy, Eren Erdal and Englund, Cristofer and Alonso-Fernandez, Fernando},
  booktitle={Proceedings of the IEEE/CVF International Conference on Computer Vision Workshops},
  pages={362--371},
  year={2023},
  doi={10.1109/ICCVW60793.2023.00043}
}

@inproceedings{najibi2017ssh,
  title={Ssh: Single stage headless face detector},
  author={Najibi, Mahyar and Samangouei, Pouya and Chellappa, Rama and Davis, Larry S},
  booktitle={Proceedings of the IEEE international conference on computer vision},
  pages={4875--4884},
  year={2017},
  doi={10.1109/ICCV.2017.522}
}

@inproceedings{hu2017finding,
  title={Finding tiny faces},
  author={Hu, Peiyun and Ramanan, Deva},
  booktitle={Proceedings of the IEEE conference on computer vision and pattern recognition},
  pages={951--959},
  year={2017},
  doi={10.1109/CVPR.2017.166}
}

@inproceedings{jamal2022multi,
  title={Multi-modal unsupervised pre-training for surgical operating room workflow analysis},
  author={Jamal, Muhammad Abdullah and Mohareri, Omid},
  booktitle={International Conference on Medical Image Computing and Computer-Assisted Intervention},
  pages={453--463},
  year={2022},
  organization={Springer},
  doi={10.1007/978-3-031-16449-1_43}
}

@article{kalman1960new,
  title={A new approach to linear filtering and prediction problems},
  author={Kalman, Rudolph Emil},
  journal={Journal of Basic Engineering},
  volume={82},
  number={1},
  pages={35--45},
  year={1960},
  publisher={American Society of Mechanical Engineers},
  doi={10.1115/1.3662552}
}

@inproceedings{wojke2017simple,
  title={Simple online and realtime tracking with a deep association metric},
  author={Wojke, Nicolai and Bewley, Alex and Paulus, Dietrich},
  booktitle={2017 IEEE international conference on image processing (ICIP)},
  pages={3645--3649},
  year={2017},
  organization={IEEE},
  doi={10.1109/ICIP.2017.8296962}
}

@article{stanojevic2024boosttrack,
  title={BoostTrack: Boosting the similarity measure and detection confidence for improved multiple object tracking},
  author={Stanojevic, Vukasin D and Todorovic, Branimir T},
  journal={Machine Vision and Applications},
  volume={35},
  number={3},
  pages={53},
  year={2024},
  publisher={Springer},
  doi={10.1007/s00138-024-01531-5}
}

@article{aharon2022bot,
  title={BoT-SORT: Robust associations multi-pedestrian tracking},
  author={Aharon, Nir and Orfaig, Roy and Bobrovsky, Ben-Zion},
  journal={arXiv preprint arXiv:2206.14651},
  year={2022},
  doi={10.48550/arXiv.2206.14651}
}

@inproceedings{maggiolino2023deep,
  title={Deep oc-sort: Multi-pedestrian tracking by adaptive re-identification},
  author={Maggiolino, Gerard and Ahmad, Adnan and Cao, Jinkun and Kitani, Kris},
  booktitle={2023 IEEE International conference on image processing (ICIP)},
  pages={3025--3029},
  year={2023},
  organization={IEEE},
  doi={10.1109/ICIP49359.2023.10222576}
}

@inproceedings{gan2021self,
  title={Self-supervised multi-view multi-human association and tracking},
  author={Gan, Yiyang and Han, Ruize and Yin, Liqiang and Feng, Wei and Wang, Song},
  booktitle={Proceedings of the 29th ACM international conference on multimedia},
  pages={282--290},
  year={2021},
  doi={10.1145/3474085.3475177}
}

@article{vo2020self,
  title={Self-supervised multi-view person association and its applications},
  author={Vo, Minh and Yumer, Ersin and Sunkavalli, Kalyan and Hadap, Sunil and Sheikh, Yaser and Narasimhan, Srinivasa G},
  journal={IEEE transactions on pattern analysis and machine intelligence},
  volume={43},
  number={8},
  pages={2794--2808},
  year={2020},
  publisher={IEEE},
  doi={10.1109/TPAMI.2020.2974726}
}

@article{luna2022graph,
  title={Graph neural networks for cross-camera data association},
  author={Luna, Elena and SanMiguel, Juan C and Mart{\'\i}nez, Jos{\'e} M and Carballeira, Pablo},
  journal={IEEE Transactions on Circuits and Systems for Video Technology},
  volume={33},
  number={2},
  pages={589--601},
  year={2022},
  publisher={IEEE},
  doi={10.1109/TCSVT.2022.3207223}
}

@inproceedings{seo2023vit,
  title={ViT-P3DE*: Vision Transformer Based Multi-Camera Instance Association with Pseudo 3D Position Embeddings.},
  author={Seo, Minseok and Lee, Hyuk-Jae and Nguyen, Xuan Truong},
  booktitle={IJCAI},
  pages={1340--1350},
  year={2023},
  doi={10.24963/ijcai.2023/149}
}

@article{kuhn1955hungarian,
  title={The Hungarian method for the assignment problem},
  author={Kuhn, Harold W},
  journal={Naval research logistics quarterly},
  volume={2},
  number={1-2},
  pages={83--97},
  year={1955},
  publisher={Wiley Online Library},
  doi={10.1002/nav.3800020109}
}

@inproceedings{huang2025deim,
  title={Deim: Detr with improved matching for fast convergence},
  author={Huang, Shihua and Lu, Zhichao and Cun, Xiaodong and Yu, Yongjun and Zhou, Xiao and Shen, Xi},
  booktitle={Proceedings of the Computer Vision and Pattern Recognition Conference},
  pages={15162--15171},
  year={2025},
  doi={10.1109/CVPR52734.2025.01412}
}

@inproceedings{loshchilov2019decoupled,
  title     = {{Decoupled Weight Decay Regularization}},
  author    = {Loshchilov, Ilya and Hutter, Frank},
  booktitle = {International Conference on Learning Representations},
  pages     = {4061--4078},
  year      = {2019},
  url       = {https://mlanthology.org/iclr/2019/loshchilov2019iclr-decoupled/}
}

@article{jiang2023rtmpose,
  title={Rtmpose: Real-time multi-person pose estimation based on mmpose},
  author={Jiang, Tao and Lu, Peng and Zhang, Li and Ma, Ningsheng and Han, Rui and Lyu, Chengqi and Li, Yining and Chen, Kai},
  journal={arXiv preprint arXiv:2303.07399},
  year={2023},
  doi={10.48550/arXiv.2303.07399}
}

@article{zhou2021learning,
  title={Learning generalisable omni-scale representations for person re-identification},
  author={Zhou, Kaiyang and Yang, Yongxin and Cavallaro, Andrea and Xiang, Tao},
  journal={IEEE transactions on pattern analysis and machine intelligence},
  volume={44},
  number={9},
  pages={5056--5069},
  year={2021},
  publisher={IEEE},
  doi={10.1109/TPAMI.2021.3069237}
}

@article{act1996health,
  title={Health insurance portability and accountability act of 1996},
  author={Act, Accountability and others},
  journal={Public law},
  volume={104},
  pages={191},
  year={1996}
}

@article{regulation2018general,
  title={General data protection regulation},
  author={Regulation, Protection},
  journal={Intouch},
  volume={25},
  pages={1--5},
  year={2018}
}

@inproceedings{carion2026sam,
  title={Sam 3: Segment anything with concepts},
  author={Carion, Nicolas and Gustafson, Laura and Hu, Yuan-Ting and Debnath, Shoubhik and Hu, Ronghang and Suris Coll-Vinent, Didac and Ryali, Chaitanya and Alwala, Kalyan Vasudev and Khedr, Haitham and Huang, Andrew and others},
  booktitle={International conference on learning representations},
  volume={2026},
  pages={138846--138923},
  year={2026}
}

@article{lei2025yolov13,
  title={Yolov13: Real-time object detection with hypergraph-enhanced adaptive visual perception},
  author={Lei, Mengqi and Li, Siqi and Wu, Yihong and Hu, Han and Zhou, You and Zheng, Xinhu and Ding, Guiguang and Du, Shaoyi and Wu, Zongze and Gao, Yue},
  journal={arXiv preprint arXiv:2506.17733},
  year={2025},
  doi={10.48550/arXiv.2506.17733}
}

@inproceedings{hao2024simplifying,
  title={Simplifying source-free domain adaptation for object detection: Effective self-training strategies and performance insights},
  author={Hao, Yan and Forest, Florent and Fink, Olga},
  booktitle={European Conference on Computer Vision},
  pages={196--213},
  year={2024},
  organization={Springer},
  doi={10.1007/978-3-031-72949-2_12}
}

\end{document}